\pdftrailerid{}
\documentclass[10pt,twocolumn,letterpaper]{article}
\usepackage[pagenumbers]{style/cvpr}
\makeatletter
\renewcommand\LenToUnit[1]{\strip@pt\dimexpr#1\relax}
\makeatother

\usepackage{booktabs}
\usepackage{colortbl}
\usepackage{amsmath,amsfonts,bm}
\usepackage{multirow}
\usepackage{microtype}
\usepackage{stfloats}

\DeclareRobustCommand{\gT}{g_{\raisebox{-0.3ex}{\ensuremath{\scriptstyle T}}}}
\newcommand{\urlnote}[2]{\footnote{\scriptsize\href{https://#2}{\nolinkurl{#1}}}}

\definecolor{TableHighlight}{HTML}{DDEBFC}
\definecolor{TableGroup}{HTML}{E7EBF1}
\definecolor{TableRule}{HTML}{D6DCE4}
\definecolor{TableMuted}{HTML}{5E6A7A}
\newcommand{\rowrule}{\noalign{\color{TableRule}\hrule height .3pt}}
\newcommand{\diagcell}[1]{\textit{\textcolor{TableMuted}{#1}}}
\newlength{\gnw}\newlength{\ggw}
\newlength{\hdw}
\def\gainsplit+#1\relax{#1}
\newcommand{\gainfmt}[1]{\ensuremath{\uparrow}\gainsplit#1\relax}
\newcommand{\gllap}[2]{\settowidth{\gnw}{\textbf{#1}}\settowidth{\ggw}{\textcolor{fgGain}{\gainfmt{#2}}}\llap{\textcolor{fgGain}{\gainfmt{#2}}\hspace{\dimexpr 2\tabcolsep+2.3em-0.5\gnw-0.167em-\ggw\relax}}}
\newcommand{\gcell}[3]{\gllap{#1}{#2}\makebox[4.6em][c]{\textbf{#3}}}
\newcommand{\gcelllast}[4]{\gllap{#1}{#2}\makebox[4.6em][c]{\textbf{#3}\rlap{\,\textcolor{fgGain}{\gainfmt{#4}}}}}
\newcommand{\gllapw}[2]{\settowidth{\gnw}{\textbf{#1}}\settowidth{\ggw}{\textcolor{fgGain}{\gainfmt{#2}}}\llap{\textcolor{fgGain}{\gainfmt{#2}}\hspace{\dimexpr 2\tabcolsep+.5\hdw-0.5\gnw-0.167em-\ggw\relax}}}
\newcommand{\oursrule}[1]{\multispan{#1}\xleaders\hbox to 3pt{\hss\vrule height .4pt width 1.8pt\hss}\hfill\cr}
\newcommand{\gllapg}[3]{\settowidth{\gnw}{\textbf{#1}}\settowidth{\ggw}{\textcolor{fgGain}{\gainfmt{#2}}}\llap{\textcolor{fgGain}{\gainfmt{#2}}\hspace{\dimexpr #3+.5\hdw-0.5\gnw-0.167em-\ggw\relax}}}

\definecolor{cvprblue}{rgb}{0.21,0.49,0.74}
\definecolor{fgGain}{HTML}{C8323C}
\usepackage[pagebackref,breaklinks,colorlinks,allcolors=cvprblue]{hyperref}
\newsavebox{\fgtabbox}
\usepackage{style/labpreprint}
\preprinttitle{Teaching Students to Stand on Their Own: Distilling Vision Foundation Models\\for Fine-Grained Few-Shot Class-Incremental Learning}
\preprintauthors{Junsu Kim\hspace{3.6em}Seungryul Baek}
\preprintaffiliation{Ulsan National Institute of Science and Technology}
\preprintplace{South Korea}
\newcommand{\paperauthors}{Junsu Kim, Seungryul Baek}

\hypersetup{pdftitle={Teaching Students to Stand on Their Own: Distilling Vision Foundation Models for Fine-Grained Few-Shot Class-Incremental Learning},pdfauthor={\paperauthors},pdfcreator={},pdfproducer={}}
\preprintabstract{Fine-grained few-shot class-incremental learning (FG-FSCIL) aims to incrementally learn visually similar new classes from limited images without catastrophic forgetting.
Prior work typically learns strong representations on base classes or leverages vision foundation models (VFMs) directly as backbones or via knowledge distillation.
However, VFM backbones incur high computational costs and do not guarantee fine-grained recognition.
Although knowledge distillation reduces inference cost, our preliminary experiments reveal that distilled students still lag substantially behind their VFM teachers, particularly on new classes.
To address this limitation, we propose STAND, a systematic VFM-to-student distillation design tailored for FG-FSCIL.
Specifically, STAND connects the teacher and student by equipping the student with a fixed copy of the teacher's classifier, aligning their outputs via projectors, and feeding both models composite images that combine different training images.
On these composite images, relation distillation matches how the teacher and the student relate each image to other images, while difference distillation matches how this relation changes between two composite images.
Finally, we freeze the student and classify new classes using length-calibrated prototypes, which restore the reduced similarity of few-shot prototypes.
Extensive evaluations demonstrate that STAND achieves state-of-the-art performance across FG-FSCIL datasets at low computational cost, while generalizing across student backbones and broader datasets.}
\begin{document}
\makepreprint

\section{Introduction}
\label{sec:main_intro}

\nocite{rebuffi2017icarl,petit2023fetril,magistri2024efc,tao2020TOPIC,zhang2021CEC,zhou2022FACT,vinyals2016matching_mini,krizhevsky2009CIFAR100,pan2024pfr,zhao2025mel,yin2026lgf,nie2025penet,song2023SAVC,park2024PriViLege,chen2025BiMC,lee2026hycal,radford2021CLIP,hinton2015distilling,oquab2023DINOv2,simeoni2025dinov3,jeong2026skd,xu2026hcd,beyer2022knowledge,zhang2018mixup}
\suppressfloats[t]
\begin{figure}[t]
\centering
\input{main/figures/teaser_links.tex}
\caption{\textbf{FG-FSCIL performance comparison.}
STAND (ours) achieves substantial gains in average accuracy on FGVC-Aircraft and Stanford Cars with ResNet-12 (scratch), even surpassing pretrained ResNet-18 and vision transformers (ViTs). Arrows give our gain over the second-best method, which is underlined.}
\label{fig:teaser}
\end{figure}

Deep neural networks recognize a fixed set of classes well, yet new classes keep appearing after training. To handle this, class-incremental learning (CIL)~\cite{rebuffi2017icarl,petit2023fetril,magistri2024efc} adds new classes over sessions while keeping earlier classes recognizable. In real-world scenarios, new classes often come with limited images, driving the emergence of few-shot class-incremental learning (FSCIL)~\cite{tao2020TOPIC,zhang2021CEC,zhou2022FACT}, where a model is first trained on base classes with sufficient images and then learns each new class from only a few images. Yet prior FSCIL methods mostly target datasets with clearly distinct classes (\eg, miniImageNet~\cite{vinyals2016matching_mini} and CIFAR-100~\cite{krizhevsky2009CIFAR100}) and fall short on fine-grained classes, which differ only in subtle visual details and are common in practice. Fine-grained FSCIL (FG-FSCIL)~\cite{pan2024pfr,zhao2025mel,yin2026lgf} thus asks a model to learn such classes from limited images. In this setting, methods~\cite{pan2024pfr,nie2025penet} typically freeze the representation learned from the base classes and represent each new class by a prototype built from the mean feature of its images. Since this base representation is not updated for new classes, FG-FSCIL needs it to already capture the subtle details that separate these classes. Accordingly, existing work strengthens it in two directions: learning more from the base images and leveraging large pretrained models.

The first direction learns more from the base images to prepare for new classes. Some methods derive virtual classes from these images to reserve room for them in the feature space~\cite{zhou2022FACT,song2023SAVC,yin2026lgf}, while others refine fine-grained features with frequency components~\cite{pan2024pfr} or masked images~\cite{nie2025penet}. Despite these additions, the effectiveness remains limited on fine-grained datasets, as shown in Fig.~\ref{fig:teaser}.

The second direction leverages large pretrained models called vision foundation models (VFMs), which carry general visual knowledge from large-scale pretraining. The most direct way to harness them is to use them as backbones, as in PriViLege~\cite{park2024PriViLege}, which tunes a pretrained vision transformer (ViT) with prompts, and in BiMC~\cite{chen2025BiMC} and HyCal~\cite{lee2026hycal}, which classify with prototypes built on frozen CLIP~\cite{radford2021CLIP} features. Yet none of them stays strong on both fine-grained datasets in Fig.~\ref{fig:teaser}, showing that a large model alone does not ensure fine-grained recognition. Moreover, it incurs a high computational cost, naturally motivating knowledge distillation (KD)~\cite{hinton2015distilling}, which transfers a large teacher's knowledge into a small student. LGF~\cite{yin2026lgf} applies it with a CLIP teacher, yet its accuracy in Fig.~\ref{fig:teaser} remains modest. This raises a question: \textit{Do VFMs hold the knowledge that FG-FSCIL needs, and if so, how much of it reaches a small student?}

To answer this, we conduct a pilot study (Sec.~\ref{sec:main_gaps}) with several VFMs. First, we train a linear classifier on their frozen features~\cite{oquab2023DINOv2,simeoni2025dinov3} with the base classes. Second, we distill each VFM into a small student using standard KD~\cite{hinton2015distilling}, recent distillation methods~\cite{jeong2026skd,xu2026hcd}, or MixUp inputs~\cite{beyer2022knowledge,zhang2018mixup}. The strongest VFMs far exceed recent FG-FSCIL methods, suggesting that they hold the knowledge that FG-FSCIL needs. However, much of it fails to reach the student. Even the best student stays well below its teacher, particularly on new classes, while also lagging on base classes. These observations motivate a more systematic distillation design that keeps the base representation useful for new classes.

In this paper, we propose STAND, a systematic design that distills VFM knowledge into a small student for FG-FSCIL by shaping the structure, inputs, and objectives of distillation together. First, we connect the two models by giving the student a copy of a classifier trained on the teacher and aligning their outputs with projectors. Next, we feed both models composite images that combine different training images, exposing the student to richer visual information than individual base classes offer. Finally, based on these inputs, we propose relation distillation and difference distillation, which teach the student how the teacher compares each input with other images and how this comparison changes between two inputs, respectively. These choices let VFM knowledge reach the student, which then stands on its own beyond the base classes. Further, since new classes have only a few images, their prototypes are less similar to test images than those of base classes, so we introduce length-calibrated prototypes, which restore this similarity and let new classes compete with base classes. Extensive experiments show that STAND achieves state-of-the-art performance on FG-FSCIL datasets, and our analyses verify its flexibility across student backbones, generality on broader datasets, and efficiency. Our contributions are as follows:
\begin{itemize}
\item We propose STAND, a systematic design that distills VFM knowledge into a small student for FG-FSCIL, keeping the base representation useful for new classes.
\item We introduce length-calibrated prototypes, which restore the reduced similarity of few-shot prototypes.
\item We demonstrate the effectiveness, flexibility, generality, and efficiency of STAND through in-depth analyses.
\end{itemize}

\section{Related work}
\label{sec:main_related}

\subsection{Continual learning (CL)}
\label{sec:main_related_cl}

\noindent\textbf{Few-shot class-incremental learning (FSCIL).}
In CIL, replay methods revisit stored examples to retain earlier classes in classification~\cite{rebuffi2017icarl} and object detection~\cite{kim2024classwise}. FSCIL extends CIL to new classes with limited images~\cite{tao2020TOPIC}. To prepare for these new classes, FACT~\cite{zhou2022FACT}, SAVC~\cite{song2023SAVC}, and DPCL~\cite{mao2026dpcl} shape the base representation with virtual classes derived from the base classes, while TEEN~\cite{wang2023teen} and ADBS~\cite{li2025adbs} adjust the classifier with calibrated prototypes or per-class decision boundaries. Fixed class directions arranged as a simplex equiangular tight frame (ETF) guide feature learning in FSCIL~\cite{yang2023NC_FSCIL} and tiny-pollinator detection~\cite{kim2026bee}. Further, since updating the representation with limited images causes overfitting and forgetting, most methods freeze their base representation while learning new classes~\cite{zhang2021CEC,zhou2022FACT,wang2023teen,oh2024CLOSER}. We follow this practice, freezing the base representation and representing new classes with their prototypes.

\noindent\textbf{Fine-grained FSCIL (FG-FSCIL).}
FG-FSCIL further requires a model to distinguish classes with subtle differences from limited images. PFR~\cite{pan2024pfr} addresses this with image frequency components, PE-Net~\cite{nie2025penet} reduces background influence with masked images, and MEL~\cite{zhao2025mel} updates its feature extractor as new classes arrive. More recently, LGF~\cite{yin2026lgf} borrows pretrained knowledge from CLIP~\cite{radford2021CLIP} through distillation for fine-grained classification.

\subsection{Vision foundation models for CL}
\label{sec:main_related_vfm}

Vision foundation models (VFMs) pretrained on images alone~\cite{oquab2023DINOv2,simeoni2025dinov3} or with text~\cite{radford2021CLIP,tschannen2025siglip2} have played three roles in CL. First, they serve as backbones for recognition. PriViLege~\cite{park2024PriViLege} tunes a pretrained ViT~\cite{dosovitskiy2021vit} with prompts and language-model knowledge; BiMC~\cite{chen2025BiMC} calibrates visual and language prototypes from CLIP~\cite{radford2021CLIP} without training; HyCal~\cite{lee2026hycal} combines cosine and Mahalanobis distances on frozen CLIP features. Second, generative VFMs such as diffusion models~\cite{Rombach_2022LDM,esser2024sd3} serve as generators that replay earlier classes~\cite{kim2024sddgr,kim2025can,wu2025dcs} and as feature backbones~\cite{kim2025beyond}. Third, VFMs supervise other models during training. VLM-PL uses a vision-language model to refine pseudo-labels for incremental object detection~\cite{kim2024vlmpl}, and LGF distills CLIP into a student for fine-grained classification~\cite{yin2026lgf}. Although effective, the first two roles incur a high cost by running a large model at inference or for generation. Thus, we study how to distill VFM knowledge into a small student for FG-FSCIL.

\subsection{Knowledge distillation}
\label{sec:main_related_kd}

Knowledge distillation (KD) transfers knowledge from a large teacher to a small student~\cite{hinton2015distilling,sariyildiz2024unic}, and prior methods vary its structure, inputs, and objectives. For the structure, SimKD~\cite{chen2022simkd} aligns the student's features to match the teacher classifier, and HCD~\cite{xu2026hcd} combines teacher and student features into shared logits across different architectures. For the inputs, FunMatch~\cite{beyer2022knowledge} feeds both models the same MixUp~\cite{zhang2018mixup} images, and \citet{marrie2024good} distill on synthetic images. For the objectives, students match the teacher's outputs~\cite{zhao2022dkd,huang2022dist}, features~\cite{romero2015fitnets}, or relations among images~\cite{park2019rkd,passalis2018pkt,koohpayegani2020compress,fang2021seed}, and SKD~\cite{jeong2026skd} matches logits and their Gram matrix. These methods are mostly developed on large datasets such as ImageNet~\cite{deng2009imagenet} and evaluated on the classes used for distillation. In contrast, FG-FSCIL needs the student to also recognize new classes that later arrive with limited images, so we design the structure, inputs, and objectives of distillation together.

\begin{figure}[t]
\centering
\includegraphics[width=\linewidth]{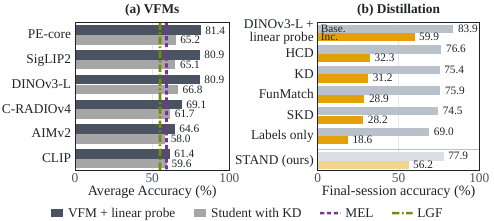}
\par\vspace{-2pt}
\caption{\textbf{VFM knowledge transfer on FGVC-Aircraft.}
(a) Average accuracy across sessions of each VFM with a linear classifier and its ResNet-12 student trained with KD.
(b) Final-session base (Base.) and new-class (Inc.) accuracy with a DINOv3-L teacher and a ResNet-12 student. Labels only means no distillation.}
\label{fig:pilot}
\end{figure}

\section{Background}
\label{sec:main_background}

\subsection{Preliminaries}
\label{sec:main_preliminaries}

\noindent\textbf{FG-FSCIL.}
FG-FSCIL starts with a base session, in which the model receives the full training set of the base classes. Then, each incremental session introduces new classes with a few images each, as in a 10-way, 5-shot session with ten new classes of five images. Each session provides only its own images, without those from earlier sessions. After each session, the model is tested on all classes seen so far.

\noindent\textbf{Prototype classifier.}
A prototype classifier compares image features with a representative vector for each class. For class $c$, a feature extractor $f$ first maps each image in its training set $\mathcal{X}_c$ to a feature normalized to unit length. Next, their mean $m_c$ is rescaled to unit length, giving the prototype $\hat{m}_c$:
\begin{equation}
\label{eq:main_prototype}
m_c=\frac{1}{n_c}\sum_{x\in\mathcal{X}_c}f(x),\qquad \hat{m}_c=\frac{m_c}{\lVert m_c\rVert},
\end{equation}
where $n_c$ is the number of images in $\mathcal{X}_c$. At inference, the predicted class $\hat{y}$ of a test image $x$ is the one whose prototype is most similar in direction, $\hat{y}=\arg\max_{c}\langle f(x),\hat{m}_c\rangle$, where the similarity $\langle\cdot,\cdot\rangle$ is the inner product.

\noindent\textbf{Knowledge distillation.}
KD~\cite{hinton2015distilling} trains a student to match a teacher's class probability distribution for the same image. Softmax converts the teacher and student class scores (logits), $o_T$ and $o_S$, into the two distributions. Then, the student minimizes their Kullback--Leibler (KL) divergence $\mathcal{L}_{\mathrm{KD}}$,
\begin{equation}
\label{eq:main_kd}
\mathcal{L}_{\mathrm{KD}}=\mathrm{KL}\bigl(\mathrm{softmax}(o_T/\tau)\,\Vert\,\mathrm{softmax}(o_S/\tau)\bigr),
\end{equation}
where a higher temperature $\tau$ gives softer distributions.

\subsection{Pilot study}
\label{sec:main_gaps}

To see whether VFMs hold the knowledge that FG-FSCIL needs, we train a linear classifier on the frozen features of VFMs~\cite{bolya2025perception,tschannen2025siglip2,simeoni2025dinov3,ranzinger2026cradio,fini2025aimv2,radford2021CLIP} with the base classes and classify new classes with prototypes. In Fig.~\ref{fig:pilot}(a), the strongest VFMs reach far higher average accuracy across sessions than the recent FG-FSCIL methods MEL~\cite{zhao2025mel} and LGF~\cite{yin2026lgf}, suggesting that their frozen features support fine-grained recognition. Then, we distill each VFM into a student with KD~\cite{hinton2015distilling} on the base classes and evaluate it in the same way to see how much of this knowledge transfers. This simple transfer already makes the students competitive with MEL and LGF, yet they stay well below their teachers, leaving room for more effective transfer.

To see whether existing distillation methods can close this gap, we compare them in Fig.~\ref{fig:pilot}(b) with DINOv3-L, the teacher of the best student. Although the KD student clearly improves over the student trained without distillation (Labels only), it reaches only about half of the teacher's final-session new-class accuracy (Inc.), while its base-class accuracy (Base.) is much closer. Even when the structure is changed with HCD~\cite{xu2026hcd}, the inputs with FunMatch~\cite{beyer2022knowledge}, or the objective with SKD~\cite{jeong2026skd}, the new-class gap to the teacher remains almost unchanged.

Therefore, knowledge distillation brings the student closer to the teacher on the base classes it is trained on, but does not ensure that VFM knowledge reaches new classes that later arrive with limited images. This calls for a more systematic design that shapes the structure, inputs, and objectives of distillation together, so that VFM knowledge also reaches classes the student has not seen. Built on such a design, STAND brings the student closest to the teacher on both base and new classes in Fig.~\ref{fig:pilot}(b).

\section{Method}
\label{sec:main_method}

\begin{figure*}[t]
\centering
\includegraphics[width=\linewidth]{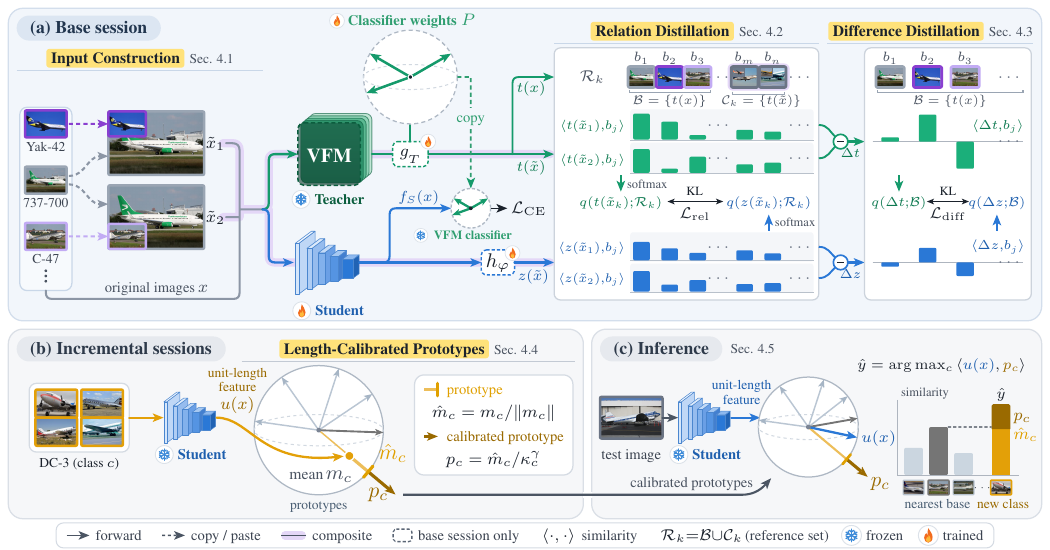}
\par\vspace{-2pt}
\caption{\textbf{Overview of STAND.}
(a) In the base session, we train the teacher projector $\gT$ and classifier weights $P$ on a frozen VFM and copy $P$ as the student's fixed VFM classifier.
Next, the teacher and the student $f_S$ with projector $h_\varphi$ take original images $x$ and their composite images $\tilde{x}_1$ and $\tilde{x}_2$. While the student classifies $x$ ($\mathcal{L}_{\mathrm{CE}}$), relation distillation ($\mathcal{L}_{\mathrm{rel}}$) matches the similarity distributions $q$ of the representations $t$ and $z$ to shared references $b_j$, and difference distillation ($\mathcal{L}_{\mathrm{diff}}$) does so for their differences $\Delta t$ and $\Delta z$.
(b) In incremental sessions, each new-class prototype $\hat{m}_c$ is built from the frozen student's unit-length features $u(x)$ and calibrated with the length ratio $\kappa_c$ and strength $\gamma$.
(c) At inference, a test image is assigned to the class $\hat{y}$ of the most similar calibrated prototype $p_c$.}
\label{fig:overview}
\end{figure*}

Our approach, STAND, sets up the distillation structure and inputs (Sec.~\ref{sec:main_teacher_student}) and defines two distillation objectives (Secs.~\ref{sec:main_relation} and~\ref{sec:main_difference}). Then, the frozen student recognizes new classes with length-calibrated prototypes (Sec.~\ref{sec:main_calibration}). Sec.~\ref{sec:main_training} gives the protocols of each session, and Fig.~\ref{fig:overview} shows a schematic overview of the whole method.

\subsection{Distillation structure and inputs}
\label{sec:main_teacher_student}

We first specialize the pretrained VFM for the base classes and transfer its classifier to the student. Specifically, we train a linear teacher projector $\gT$ and classifier weights $P=\{P_c\}$ on top of its frozen feature extractor $f_T$ using base-class labels, where $P_c\in\mathbb{R}^{d_T}$ is a unit-length weight vector for each base class $c$. Then, for the student to distinguish the base classes as the teacher does, we freeze both and transfer these weights to the student as a fixed classifier, rather than training a separate one. Since the student feature extractor $f_S$ outputs features $f_S(x)\in\mathbb{R}^{d_S}$ of a different dimension, we copy each weight vector as $QP_c\in\mathbb{R}^{d_S}$ using a fixed orthonormal map $Q$, preserving the angles between them. The copied weights form the fixed VFM classifier, with which we train the student on the original training images.

Along with transferring the classifier, we let the student learn the teacher's responses to input images, which requires comparing the outputs of the two models in one space. Thus, we map the student features $f_S(x)$ into the output space of the teacher projector $\gT$ with a trainable student projector $h_\varphi:\mathbb{R}^{d_S}\to\mathbb{R}^{d_T}$. This gives the teacher representation $t(x)$ and the student representation $z(x)$ of an image $x$:
\begin{equation}
\label{eq:main_representations}
t(x)=\nu\bigl(\gT(f_T(x))\bigr),\quad
z(x)=\nu\bigl(h_\varphi(f_S(x))\bigr),
\end{equation}
where $\nu(\cdot)$ normalizes a vector to unit length.

\noindent\textbf{Input construction.}
As the inputs for these responses, we construct composite images that combine different training images, exposing the student to richer visual information than individual base-class images offer. Specifically, given an original image $x$ as the background, we randomly sample two images $x_1$ and $x_2$ from the same batch. Each sampled image is independently resized and pasted at a random location onto a separate copy of $x$, producing two composite images, $\tilde{x}_1$ and $\tilde{x}_2$. The original images train the student with the fixed VFM classifier, while the composite images are fed to both models for distillation.

\subsection{Relation distillation}
\label{sec:main_relation}

The composite images carry visual information beyond individual base classes, which a class label cannot express. To transfer this information, we distill how the teacher relates each composite image to other images. These other images are the original images of the base session and the composite images of the current batch, which serve as references shared by both models. Specifically, the teacher representations $t(x)$ of the original images form a set $\mathcal{B}$, while those of the composite images other than $\tilde{x}_k$ form a set $\mathcal{C}_k$. Together, they define the reference set $\mathcal{R}_k=\mathcal{B}\cup\mathcal{C}_k$ of references $b_j$.

For each composite image $\tilde{x}_k$, the teacher and the student each compare their own representation with the same references in $\mathcal{R}_k$. Next, a softmax with temperature $\tau$, as in Eq.~\ref{eq:main_kd}, converts the resulting similarities into a distribution:
\begin{equation}
\label{eq:main_relation_distribution}
q_j(v;\mathcal{R}_k)=
\frac{\exp(\langle v,b_j\rangle/\tau)}
{\sum_{b_l\in\mathcal{R}_k}\exp(\langle v,b_l\rangle/\tau)},
\end{equation}
where $v$ denotes either the teacher representation $t(\tilde{x}_k)$ or the student representation $z(\tilde{x}_k)$. We refer to $q(v;\mathcal{R}_k)$ as the response of $v$. By matching these responses, the student learns to compare each composite image with the references as the teacher does. The relation distillation loss is the KL divergence between the teacher and student responses, averaged over the composite images in the batch:
\begin{equation}
\label{eq:main_relation_loss}
\mathcal{L}_{\mathrm{rel}}=
\mathbb{E}_{k}\,
\mathrm{KL}\bigl(q(t(\tilde{x}_k);\mathcal{R}_k)\,\Vert\,q(z(\tilde{x}_k);\mathcal{R}_k)\bigr).
\end{equation}

\subsection{Difference distillation}
\label{sec:main_difference}

The two composite images of $x$ share its background and differ only in what is pasted, so the change in the teacher's response between them reflects this difference. Relation distillation conveys this change only weakly, since KL divergence places little emphasis on references with low teacher probabilities. To capture this change, difference distillation explicitly transfers how the teacher's response varies between the two composite images, using response-difference supervision~\cite{liu2024dccd} over the references in $\mathcal{B}$, which both composite images share. Specifically, we compute the representation differences $\Delta t=t(\tilde{x}_1)-t(\tilde{x}_2)$ and $\Delta z=z(\tilde{x}_1)-z(\tilde{x}_2)$. Their inner products with each reference $b_j$ measure the change in its similarity, corresponding to the relative change in its probability. Next, we compute the responses of $\Delta t$ and $\Delta z$ using Eq.~\ref{eq:main_relation_distribution}, with $\mathcal{B}$ replacing $\mathcal{R}_k$. The response of $\Delta t$ emphasizes references whose similarities increase most from $\tilde{x}_2$ to $\tilde{x}_1$, while that of $-\Delta t$ emphasizes those with the largest decreases. Therefore, we distill both directions using a sign $\sigma$, yielding the difference distillation loss:
\begin{equation}
\label{eq:main_difference_loss}
\mathcal{L}_{\mathrm{diff}}=
\mathbb{E}_{\sigma\in\{-1,1\}}\,
\mathrm{KL}\bigl(q(\sigma\Delta t;\mathcal{B})\,\Vert\,q(\sigma\Delta z;\mathcal{B})\bigr).
\end{equation}

\subsection{Length-calibrated prototypes}
\label{sec:main_calibration}

Once distillation ends, we freeze the student and build class prototypes with Eq.~\ref{eq:main_prototype} from unit-length features $u(x)$ obtained from its features $f_S(x)$. However, a prototype built from only a few images is less similar to test images of its class than one built from many, so new classes tend to lose to base classes. This lower similarity arises because image-specific variations do not fully cancel out in the mean $m_c$ of a few images. The remaining variations add to its length, so $m_c$ tends to be longer than the population mean $\mu_c$, the expected feature of class $c$. This length difference reduces the similarity between test features and the prototype $\hat{m}_c$. Specifically, for a test feature of class $c$, the expected inner product with $m_c$ is $\lVert\mu_c\rVert^2$, regardless of the number of images. However, normalization divides this inner product by $\lVert m_c\rVert$ rather than $\lVert\mu_c\rVert$, reducing the expected similarity to approximately $\lVert\mu_c\rVert/\lVert m_c\rVert$ of that obtained with the population prototype $\mu_c/\lVert\mu_c\rVert$.

Since $\mu_c$ is unknown, we estimate this fraction from the images of class $c$ using the length ratio
\begin{equation}
\label{eq:main_lcp_ratio}
\kappa_c=\frac{\sqrt{\widehat{\rho}_c}}{\lVert m_c\rVert},
\end{equation}
where $\widehat{\rho}_c$ estimates $\lVert\mu_c\rVert^2$ as the mean similarity between distinct images of class $c$. As the number of images increases, $\kappa_c$ approaches one, so base classes keep nearly all of their similarity, whereas new classes do not. Since the similarity is an inner product, which grows with prototype length, we restore the reduced similarity by scaling $\hat{m}_c$ by $\kappa_c^{-\gamma}$, giving length-calibrated prototypes (LCP) $p_c$:
\begin{equation}
\label{eq:main_lcp}
p_c=\frac{\hat{m}_c}{\kappa_c^\gamma},
\end{equation}
where the LCP strength $\gamma$ controls how much of the reduced similarity is restored. This calibration lets new classes compete with base classes, as in Fig.~\ref{fig:overview}(c). Supp.~Mat.\ gives the derivation and measured values of $\kappa_c$.

\subsection{Protocols}
\label{sec:main_training}

\noindent\textbf{Base session.}
We first train $\gT$ and $P$ with cross-entropy loss and form the fixed VFM classifier as in Sec.~\ref{sec:main_teacher_student}. Then, we train the student feature extractor $f_S$ and projector $h_\varphi$ with cross-entropy loss $\mathcal{L}_{\mathrm{CE}}$ on original images and the two distillation losses on composite images. The total loss is
\begin{equation}
\label{eq:main_total_loss}
\mathcal{L}=\mathcal{L}_{\mathrm{CE}}
+\lambda_r\mathcal{L}_{\mathrm{rel}}
+\lambda_d\mathcal{L}_{\mathrm{diff}}.
\end{equation}
Here, $\lambda_r$ and $\lambda_d$ weight relation and difference distillation, respectively. After base-session training, we freeze $f_S$ and discard the teacher, the fixed VFM classifier, and the student projector. From the student's unit-length features $u(x)$, we construct length-calibrated prototypes $p_c$ for the base classes.

\noindent\textbf{Incremental sessions and inference.}
In each incremental session, we add length-calibrated prototypes for the new classes, built from their limited images, without updating the student. At inference, we classify each test image $x$ as the class whose calibrated prototype has the highest inner product with its feature $u(x)$, $\hat{y}=\arg\max_c\langle u(x),p_c\rangle$.

\section{Experiments}
\label{sec:main_experiments}

\begin{figure*}[t]
\captionof{table}{\textbf{Comparison with FSCIL and FG-FSCIL methods on four datasets.} All methods use ResNet-12 (scratch). Our results are averaged over three seeds. Acc./AA.: final-session/average accuracy (\%). \textbf{Bold}/\underline{underline}/-: best/second-best/unavailable. $\uparrow$: our gain over the second-best.}
\label{tab:comparison}
\centering
\setlength{\tabcolsep}{3pt}
\renewcommand{\arraystretch}{1.2}
\scriptsize
\sbox{\fgtabbox}{\begin{tabular}{lllcccccccc@{\hspace{1.8em}}}
\toprule
\multirow{2}{*}[-2.4pt]{\footnotesize\textbf{Method}} & \multicolumn{1}{l}{\multirow{2}{*}[-2.4pt]{\footnotesize\textbf{Venue}}} & \multicolumn{1}{l}{\multirow{2}{*}[-2.4pt]{\footnotesize\textbf{Task}}} & \multicolumn{2}{c}{\footnotesize\textbf{FGVC-Aircraft}} & \multicolumn{2}{c}{\footnotesize\textbf{Stanford Cars}} & \multicolumn{2}{c}{\footnotesize\textbf{CUB-200}} & \multicolumn{2}{c@{\hspace{1.8em}}}{\footnotesize\textbf{Stanford Dogs}} \\
\cmidrule(lr){4-5} \cmidrule(lr){6-7} \cmidrule(lr){8-9} \cmidrule(lr{14.6pt}){10-11}
 &  &  & \footnotesize Acc. & \footnotesize AA. & \footnotesize Acc. & \footnotesize AA. & \footnotesize Acc. & \footnotesize AA. & \footnotesize Acc. & \footnotesize AA. \\
\midrule
FACT~\cite{zhou2022FACT} & CVPR'22 & FSCIL & 31.60 & 40.46 & 43.98 & 56.17 & 32.71 & 42.50 & 33.26 & 42.96 \\
\rowrule
TEEN~\cite{wang2023teen} & NeurIPS'23 & FSCIL & 30.98 & 43.42 & 45.32 & 57.37 & 30.08 & 39.33 & 39.14 & 47.86 \\
\rowrule
CLOSER~\cite{oh2024CLOSER} & ECCV'24 & FSCIL & 36.56 & 48.08 & 34.49 & 53.78 & 31.74 & 43.95 & 34.37 & 44.69 \\
\rowrule
ADBS~\cite{li2025adbs} & AAAI'25 & FSCIL & 35.42 & 45.23 & 29.79 & 46.50 & 27.85 & 38.71 & 29.15 & 41.54 \\
\rowrule
DPCL~\cite{mao2026dpcl} & CVPRF'26 & FSCIL & 46.12 & 55.72 & \underline{57.61} & \underline{69.80} & 45.04 & 55.74 & 34.50 & 43.54 \\
\rowrule
PFR~\cite{pan2024pfr} & PR'24 & FG-FSCIL & 34.42 & 43.05 & 44.56 & 58.12 & 33.99 & 43.70 & 36.01 & 44.25 \\
\rowrule
PE-Net~\cite{nie2025penet} & ICONIP'25 & FG-FSCIL & 36.56 & 49.53 & 46.18 & 60.12 & - & - & 38.76 & 47.34 \\
\rowrule
MEL~\cite{zhao2025mel} & NeurIPS'25 & FG-FSCIL & \underline{50.06} & \underline{59.23} & 57.58 & 68.43 & 36.00 & 45.44 & \underline{44.06} & \underline{52.38} \\
\rowrule
LGF~\cite{yin2026lgf} & IJMIR'26 & FG-FSCIL & 46.28 & 55.07 & 50.83 & 62.95 & \underline{54.46} & \underline{64.61} & 42.74 & 52.30 \\
\oursrule{11}
\rowcolor{TableHighlight}[\tabcolsep][1.8em] STAND (ours) &  & FG-FSCIL & \makebox[4.6em][c]{\textbf{69.23}} & \gcell{69.23}{+19.2}{76.60} & \gcell{76.60}{+17.4}{78.26} & \gcell{78.26}{+20.7}{85.57} & \gcell{85.57}{+15.8}{65.57} & \gcell{65.57}{+11.1}{72.44} & \gcell{72.44}{+7.8}{52.76} & \gcelllast{52.76}{+8.7}{61.74}{+9.4} \\
\bottomrule
\end{tabular}}
\ifdim\wd\fgtabbox>\linewidth
  \resizebox{\linewidth}{!}{\usebox{\fgtabbox}}\else
  \usebox{\fgtabbox}\fi

\par\vspace{6pt}\noindent\normalsize
\vspace{5pt}
\includegraphics[width=\linewidth]{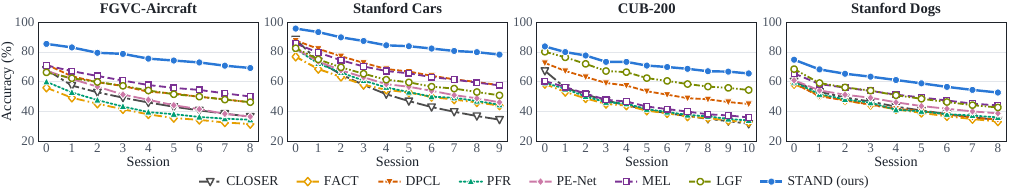}
\captionsetup{position=top}\captionof{figure}{\textbf{Comparison with FSCIL and FG-FSCIL methods across sessions.} All methods use ResNet-12 (scratch) on four datasets. Our results are averaged over three seeds. Accuracy (\%) is measured over all classes seen up to each session.}
\label{fig:sessions}
\vspace{-5pt}
\end{figure*}

\subsection{Experimental setup}
\label{sec:main_implementation}

\noindent\textbf{Implementation details.}
We follow the benchmark protocol of prior work~\cite{pan2024pfr,zhao2025mel}. FGVC-Aircraft~\cite{maji2013fgvc} has 60 base classes followed by 8 incremental sessions (5-way, 5-shot); Stanford Cars~\cite{krause2013cars} has 106 followed by 9 (10-way, 5-shot); CUB-200~\cite{wah2011CUB200} has 100 followed by 10 (10-way, 5-shot); and Stanford Dogs~\cite{khosla2011dogs} has 80 followed by 8 (5-way, 5-shot). We further use CIFAR-100~\cite{krizhevsky2009CIFAR100} and miniImageNet~\cite{vinyals2016matching_mini}, each with 60 base classes followed by 8 incremental sessions (5-way, 5-shot), as well as EMDS-7~\cite{yang2023emds7}, with 19 followed by 5 (3-way, 3-shot). Unless stated otherwise, the teacher is a frozen DINOv3-L~\cite{simeoni2025dinov3}, and the student is a ResNet-12 trained from scratch. More details are in Supp.~Mat.

\noindent\textbf{Evaluation metrics.}
After each session, we measure accuracy on all seen classes. AA.\ averages it over sessions, and Acc.\ is its final-session value, which we also report on base and new classes (Base.\ and Inc.). Differences are in percentage points (pp). For all metrics, higher is better.

\subsection{Comparison with existing methods}
\label{sec:main_comparison}

Tab.~\ref{tab:comparison} and Fig.~\ref{fig:sessions} compare STAND (ours) with FSCIL and FG-FSCIL methods on the same ResNet-12 (scratch). Ours reaches the best accuracy on all four datasets and leads in every session (details in Supp.~Mat.). Notably, its advantage over existing methods grows toward the final session, indicating that the learned representation remains effective as new classes arrive. In particular, FSCIL methods such as FACT~\cite{zhou2022FACT} and CLOSER~\cite{oh2024CLOSER} prepare for new classes with virtual classes or contrastive learning, but their accuracy drops steeply as these classes arrive. On Stanford Cars, even CLOSER drops from 88.56\% in the base session to 34.49\% in the final one. Meanwhile, FG-FSCIL methods such as PFR~\cite{pan2024pfr} and PE-Net~\cite{nie2025penet} refine the base representation with frequency components or masked images, yet remain far behind; the second-best MEL~\cite{zhao2025mel} is still 19.2pp below ours on FGVC-Aircraft. Similar to our approach, LGF~\cite{yin2026lgf} distills a VFM (\ie, CLIP~\cite{radford2021CLIP}) into a small student, but it falls well short of ours. Neither a strong start, a refined base representation, nor a VFM teacher alone keeps accuracy high as new classes arrive. This suggests that the key is how VFM knowledge is distilled into the student, which decides whether the base representation stays useful for new classes.

\begin{table}[t]
\caption{\textbf{Component ablation.} Best seed by AA. Fixed: fixed VFM classifier; Rel./Diff.: relation/difference distillation.}
\label{tab:components}
\centering
\setlength{\tabcolsep}{2pt}
\renewcommand{\arraystretch}{1.2}
\scriptsize
\settowidth{\hdw}{\textbf{Base.}}
\sbox{\fgtabbox}{\begin{tabular}{cccc@{\hspace{2pt}}c@{\hspace{2pt}}cccc@{\hspace{3pt}}c@{\hspace{3pt}}cccc}
\toprule
\multirow{2}{*}[-2.7pt]{\textbf{Fixed}} & \multirow{2}{*}[-2.7pt]{\textbf{Rel.}} & \multirow{2}{*}[-2.7pt]{\textbf{Diff.}} & \multirow{2}{*}[-2.7pt]{\textbf{LCP}} & & \multicolumn{4}{c@{\hspace{5pt}}}{\textbf{FGVC-Aircraft}} & & \multicolumn{4}{c@{\hspace{4pt}}}{\textbf{Stanford Cars}} \\
\cmidrule(l{2pt}r{5pt}){6-9} \cmidrule(l{2pt}r{4pt}){11-14}
 &  &  &  &  & \makebox[\hdw][c]{Base.} & \makebox[\hdw][c]{Inc.} & \makebox[\hdw][c]{Acc.} & \makebox[\hdw][c]{AA.} &  & \makebox[\hdw][c]{Base.} & \makebox[\hdw][c]{Inc.} & \makebox[\hdw][c]{Acc.} & \makebox[\hdw][c]{AA.} \\
\midrule
- & - & - & - &  & 69.13 & 15.80 & 47.80 & 57.69 &  & 82.54 & 29.06 & 50.05 & 62.59 \\
\rowrule
\ensuremath{\checkmark} & - & - & - &  & 69.03 & 18.60 & 48.86 & 57.93 &  & 82.89 & 33.62 & 52.96 & 65.09 \\
\rowrule
\ensuremath{\checkmark} & \ensuremath{\checkmark} & - & - &  & 81.40 & 39.30 & 64.56 & 73.36 &  & 92.77 & 59.10 & 72.32 & 81.10 \\
\rowrule
\ensuremath{\checkmark} & - & \ensuremath{\checkmark} & - &  & 79.23 & 34.35 & 61.28 & 70.39 &  & 92.53 & 56.45 & 70.61 & 79.89 \\
\rowrule
\ensuremath{\checkmark} & \ensuremath{\checkmark} & \ensuremath{\checkmark} & - &  & \textbf{83.17} & 42.25 & 66.80 & 75.40 &  & \textbf{93.66} & 64.31 & 75.83 & 83.60 \\
\addlinespace[2pt]
\rowcolor{TableHighlight} \ensuremath{\checkmark} & \ensuremath{\checkmark} & \ensuremath{\checkmark} & \ensuremath{\checkmark} &  & 78.03 & \textbf{56.85} & \textbf{69.56} & \textbf{77.02} &  & 91.76 & \textbf{70.87} & \textbf{79.07} & \textbf{85.92} \\
\bottomrule
\end{tabular}}
\ifdim\wd\fgtabbox>\linewidth
  \resizebox{\linewidth}{!}{\usebox{\fgtabbox}}\else
  \usebox{\fgtabbox}\fi
\end{table}

\begin{table}[t]
\begingroup
\renewenvironment{table}[1][]{}{\par}
\begin{table}[t]
\caption{\textbf{\boldmath Relation distillation weight \ensuremath{\lambda}\textsubscript{r}.} Best seed by AA.}
\label{tab:relation_weight}
\centering
\setlength{\tabcolsep}{3pt}
\renewcommand{\arraystretch}{1.2}
\scriptsize
\settowidth{\hdw}{\footnotesize\textbf{Base.}}
\sbox{\fgtabbox}{\begin{tabular}{ccccccccc}
\toprule
\multirow{2}{*}[-2.4pt]{\footnotesize\textbf{\boldmath \ensuremath{\lambda}\textsubscript{r}}} & \multicolumn{4}{c}{\footnotesize\textbf{FGVC-Aircraft}} & \multicolumn{4}{c}{\footnotesize\textbf{Stanford Cars}} \\
\cmidrule(lr){2-5} \cmidrule(lr){6-9}
 & \makebox[\hdw][c]{\footnotesize Base.} & \makebox[\hdw][c]{\footnotesize Inc.} & \makebox[\hdw][c]{\footnotesize Acc.} & \makebox[\hdw][c]{\footnotesize AA.} & \makebox[\hdw][c]{\footnotesize Base.} & \makebox[\hdw][c]{\footnotesize Inc.} & \makebox[\hdw][c]{\footnotesize Acc.} & \makebox[\hdw][c]{\footnotesize AA.} \\
\midrule
0 & 74.00 & 50.25 & 64.50 & 72.32 & 90.05 & 64.10 & 74.28 & 82.49 \\
\rowrule
0.25 & 75.90 & 52.50 & 66.54 & 74.51 & 90.85 & 67.03 & 76.38 & 84.22 \\
\rowrule
0.5 & 76.73 & 52.90 & 67.20 & 75.30 & 91.87 & 69.15 & 78.07 & 85.25 \\
\rowrule
0.75 & 77.73 & 53.10 & 67.88 & 75.38 & \textbf{92.02} & 70.34 & 78.85 & 85.78 \\
\addlinespace[2pt]
\rowcolor{TableHighlight} 1 & \textbf{78.03} & \textbf{56.85} & \textbf{69.56} & \textbf{77.02} & 91.76 & \textbf{70.87} & \textbf{79.07} & \textbf{85.92} \\
\bottomrule
\end{tabular}}
\ifdim\wd\fgtabbox>\linewidth
  \resizebox{\linewidth}{!}{\usebox{\fgtabbox}}\else
  \usebox{\fgtabbox}\fi
\end{table}

\par\vspace{5pt}
\begin{table}[t]
\caption{\textbf{\boldmath Difference distillation weight \ensuremath{\lambda}\textsubscript{d}.} Best seed by AA.}
\label{tab:difference_weight}
\centering
\setlength{\tabcolsep}{3pt}
\renewcommand{\arraystretch}{1.2}
\scriptsize
\settowidth{\hdw}{\footnotesize\textbf{Base.}}
\sbox{\fgtabbox}{\begin{tabular}{ccccccccc}
\toprule
\multirow{2}{*}[-2.4pt]{\footnotesize\textbf{\boldmath \ensuremath{\lambda}\textsubscript{d}}} & \multicolumn{4}{c}{\footnotesize\textbf{FGVC-Aircraft}} & \multicolumn{4}{c}{\footnotesize\textbf{Stanford Cars}} \\
\cmidrule(lr){2-5} \cmidrule(lr){6-9}
 & \makebox[\hdw][c]{\footnotesize Base.} & \makebox[\hdw][c]{\footnotesize Inc.} & \makebox[\hdw][c]{\footnotesize Acc.} & \makebox[\hdw][c]{\footnotesize AA.} & \makebox[\hdw][c]{\footnotesize Base.} & \makebox[\hdw][c]{\footnotesize Inc.} & \makebox[\hdw][c]{\footnotesize Acc.} & \makebox[\hdw][c]{\footnotesize AA.} \\
\midrule
0 & 76.60 & 53.95 & 67.54 & 75.15 & 90.25 & 67.18 & 76.23 & 83.91 \\
\rowrule
0.25 & 76.43 & 55.25 & 67.96 & 75.55 & 91.22 & 67.15 & 76.60 & 84.34 \\
\rowrule
0.5 & 77.53 & 55.90 & 68.88 & 75.89 & 91.47 & 69.10 & 77.88 & 85.08 \\
\rowrule
0.75 & 77.00 & 56.45 & 68.78 & 76.34 & 91.60 & 69.35 & 78.09 & 85.49 \\
\addlinespace[2pt]
\rowcolor{TableHighlight} 1 & \textbf{78.03} & \textbf{56.85} & \textbf{69.56} & \textbf{77.02} & \textbf{91.76} & \textbf{70.87} & \textbf{79.07} & \textbf{85.92} \\
\bottomrule
\end{tabular}}
\ifdim\wd\fgtabbox>\linewidth
  \resizebox{\linewidth}{!}{\usebox{\fgtabbox}}\else
  \usebox{\fgtabbox}\fi
\end{table}

\endgroup
\end{table}

\subsection{Ablation study}
\label{sec:main_ablation}

\noindent\textbf{Main components.}
Tab.~\ref{tab:components} adds our components one at a time. First, the fixed VFM classifier improves AA.\ by 0.24pp on FGVC-Aircraft and 2.50pp on Stanford Cars. Relation distillation then provides the largest gain, improving both Base.\ and Inc.\ and reaching 73.36\% and 81.10\% AA., respectively. Difference distillation also improves both metrics when used alone, reaching 70.39\% and 79.89\% AA. Combining the two yields higher accuracy than either alone across all metrics, suggesting that how the teacher relates each composite image to other images and how this relation changes between composite images provide complementary knowledge. Finally, LCP substantially improves Inc., reaching the highest Acc.\ and AA. Overall, each component improves Acc.\ and AA.\ on both datasets, while their combined effect brings the student closest to its teacher in Fig.~\ref{fig:pilot}(b).

\begin{figure*}[t]
\centering
\includegraphics[width=\linewidth]{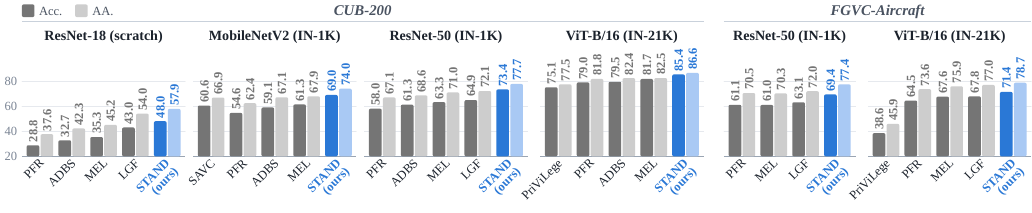}
\par\vspace{-8pt}
\caption{\textbf{Backbone flexibility.} IN-1K/IN-21K: pretrained on ImageNet-1K/21K.}
\label{fig:backbones}
\end{figure*}

\noindent\textbf{Distillation weights.}
Tabs.~\ref{tab:relation_weight} and~\ref{tab:difference_weight} vary the weight of each distillation loss in Eq.~\ref{eq:main_total_loss}, keeping the other fixed at one. For both relation and difference distillation, AA.\ increases as the weight grows and peaks at one, which we adopt throughout. Moreover, both improve Base.\ and Inc., with larger gains on Inc., showing that the knowledge they transfer goes beyond the base classes used for distillation and reaches new classes that the student has never seen.

\noindent\textbf{LCP strength.}
Tab.~\ref{tab:lcp_strength} varies the LCP strength $\gamma$, which controls how much of the reduced similarity of new-class prototypes is restored. Unlike the distillation weights, $\gamma$ shifts accuracy from base to new classes by lengthening new-class prototypes more strongly. The gap between Base.\ and Inc.\ narrows from 40.92pp at $\gamma=0$ to 21.18pp at $\gamma=0.5$ on FGVC-Aircraft. Up to $\gamma=0.5$, the gain in Inc.\ is approximately three times the change in Base.\ on both datasets. Beyond this value, the two changes become comparable, so we adopt it as the balance point. Thus, LCP lets new classes compete with base classes.

\begin{table}[t]
\caption{\textbf{\boldmath LCP strength \ensuremath{\gamma}.} Best seed by AA.}
\label{tab:lcp_strength}
\centering
\setlength{\tabcolsep}{3pt}
\renewcommand{\arraystretch}{1.2}
\scriptsize
\settowidth{\hdw}{\footnotesize\textbf{Base.}}
\sbox{\fgtabbox}{\begin{tabular}{ccccccccc}
\toprule
\multirow{2}{*}[-2.4pt]{\footnotesize\textbf{\boldmath \ensuremath{\gamma}}} & \multicolumn{4}{c}{\footnotesize\textbf{FGVC-Aircraft}} & \multicolumn{4}{c}{\footnotesize\textbf{Stanford Cars}} \\
\cmidrule(lr){2-5} \cmidrule(lr){6-9}
 & \makebox[\hdw][c]{\footnotesize Base.} & \makebox[\hdw][c]{\footnotesize Inc.} & \makebox[\hdw][c]{\footnotesize Acc.} & \makebox[\hdw][c]{\footnotesize AA.} & \makebox[\hdw][c]{\footnotesize Base.} & \makebox[\hdw][c]{\footnotesize Inc.} & \makebox[\hdw][c]{\footnotesize Acc.} & \makebox[\hdw][c]{\footnotesize AA.} \\
\midrule
0 & \textbf{83.17} & 42.25 & 66.80 & 75.40 & \textbf{93.66} & 64.31 & 75.83 & 83.60 \\
\rowrule
0.25 & 80.93 & 49.50 & 68.36 & 76.37 & 92.97 & 67.68 & 77.61 & 84.93 \\
\addlinespace[2pt]
\rowcolor{TableHighlight} 0.5 & 78.03 & 56.85 & \textbf{69.56} & 77.02 & 91.76 & 70.87 & 79.07 & 85.92 \\
\rowrule
0.75 & 73.00 & 64.20 & 69.48 & \textbf{77.03} & 89.47 & 73.35 & 79.68 & 86.48 \\
\rowrule
1 & 67.73 & \textbf{68.45} & 68.02 & 75.93 & 87.92 & \textbf{74.80} & \textbf{79.95} & \textbf{86.71} \\
\bottomrule
\end{tabular}}
\ifdim\wd\fgtabbox>\linewidth
  \resizebox{\linewidth}{!}{\usebox{\fgtabbox}}\else
  \usebox{\fgtabbox}\fi
\end{table}

\subsection{Analysis}
\label{sec:main_analysis}

\noindent\textbf{Flexibility.}
Fig.~\ref{fig:backbones} evaluates STAND (ours) with different student backbones, from ResNet-18 to ViT-B/16. Ours achieves the highest Acc.\ and AA.\ with every backbone, ahead of the FG-FSCIL methods MEL~\cite{zhao2025mel} and LGF~\cite{yin2026lgf}, which also distills a VFM. Our margin in Acc.\ over the second-best method ranges from 3.6pp with ViT-B/16 to 8.5pp with ResNet-50. With ResNet-18 trained from scratch, where the student has no pretrained knowledge, ours exceeds MEL by 12.6pp. These results show that our distillation benefits students across backbones, so the student can be chosen independently of the teacher.

\begin{table}[t]
\caption{\textbf{Comparison on standard FSCIL datasets.}}
\label{tab:standard_fscil}
\centering
\setlength{\tabcolsep}{2pt}
\renewcommand{\arraystretch}{1.2}
\scriptsize
\setlength{\hdw}{30pt}
\sbox{\fgtabbox}{\begin{tabular}{llcccc@{\hspace{1.2em}}}
\toprule
\multirow{2}{*}[-2.4pt]{\footnotesize\textbf{Method}} & \multicolumn{1}{l}{\multirow{2}{*}[-2.4pt]{\footnotesize\textbf{Backbone}}} & \multicolumn{2}{c}{\footnotesize\textbf{CIFAR-100}} & \multicolumn{2}{c@{\hspace{1.2em}}}{\footnotesize\textbf{miniImageNet}} \\
\cmidrule(lr){3-4} \cmidrule(lr{11.4pt}){5-6}
 &  & \makebox[\hdw][c]{\footnotesize Acc.} & \makebox[\hdw][c]{\footnotesize AA.} & \makebox[\hdw][c]{\footnotesize Acc.} & \makebox[\hdw][c]{\footnotesize AA.} \\
\midrule
ADBS~\cite{li2025adbs} & \diagcell{ResNet-12} & 51.15 & 63.77 & 55.38 & 66.02 \\
\rowrule
DPCL~\cite{mao2026dpcl} & \diagcell{ResNet-12} & \underline{56.54} & \underline{68.29} & \underline{58.47} & \underline{69.11} \\
\rowrule
PFR~\cite{pan2024pfr} & \diagcell{ResNet-12} & 53.21 & 64.60 & 52.61 & 63.76 \\
\rowrule
MEL~\cite{zhao2025mel} & \diagcell{ResNet-12} & 55.65 & 65.48 & 57.92 & 67.85 \\
\oursrule{6}
\rowcolor{TableHighlight}[\tabcolsep][1.2em] STAND (ours) & \diagcell{ResNet-12} & \makebox[\hdw][c]{\textbf{62.48}} & \gllapw{62.48}{+5.9}\makebox[\hdw][c]{\textbf{72.08}} & \gllapw{72.08}{+3.8}\makebox[\hdw][c]{\textbf{63.37}} & \gllapw{63.37}{+4.9}\makebox[\hdw][c]{\textbf{72.54}\rlap{\,\textcolor{fgGain}{\gainfmt{+3.4}}}} \\
\bottomrule
\end{tabular}}
\ifdim\wd\fgtabbox>\linewidth
  \resizebox{\linewidth}{!}{\usebox{\fgtabbox}}\else
  \usebox{\fgtabbox}\fi
\end{table}

\begin{figure*}[t]
\centering
\includegraphics[width=\linewidth]{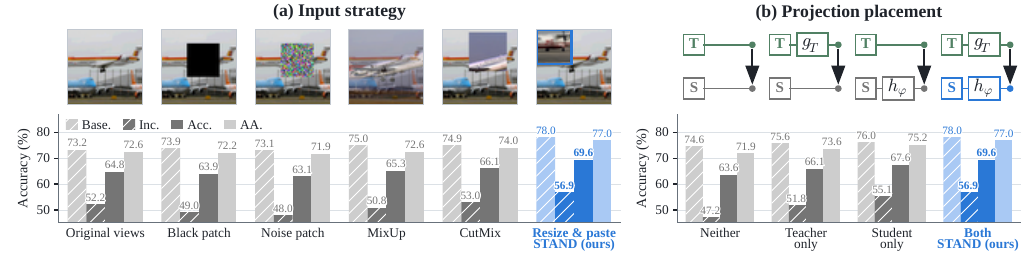}
\par\vspace{-4pt}
\caption{\textbf{Input strategy and projection placement on FGVC-Aircraft.} (a) Accuracy of each input strategy. (b) Accuracy with and without the teacher projector $\gT$ and the student projector $h_\varphi$. T/S: teacher/student.}
\label{fig:design_analysis}
\end{figure*}

\noindent\textbf{Generality.}
Tab.~\ref{tab:standard_fscil} evaluates STAND (ours) beyond fine-grained datasets, on CIFAR-100 and miniImageNet, whose classes are clearly distinct. Although tailored for fine-grained classes, ours leads on both with ResNet-12, reaching 72.08\% and 72.54\% AA. It exceeds the second-best DPCL~\cite{mao2026dpcl} by 5.9pp and 4.9pp in Acc., and leads the FG-FSCIL methods PFR~\cite{pan2024pfr} and MEL~\cite{zhao2025mel} by larger margins. Tab.~\ref{tab:emds7} further evaluates ours on EMDS-7~\cite{yang2023emds7}, a specialized dataset of microscopic images that remains challenging for FSCIL methods developed on standard datasets~\cite{zhou2026emds7}. Ours again achieves the highest Acc., leading by 11.5pp with ResNet-12 and 5.3pp with ResNet-18. These results suggest that STAND is not restricted to particular datasets and can be applied across a broad range of domains.

\begin{table}[t]
\caption{\textbf{Comparison on EMDS-7.}}
\label{tab:emds7}
\centering
\setlength{\tabcolsep}{2pt}
\renewcommand{\arraystretch}{1.2}
\scriptsize
\setlength{\hdw}{33pt}
\sbox{\fgtabbox}{\begin{tabular}{lcc@{\hspace{1.2em}}>{\hspace{8pt}}lcc@{\hspace{1.2em}}}
\toprule
\multirow{2}{*}[-2.4pt]{\footnotesize\textbf{Method}} & \multicolumn{2}{c@{\hspace{1.2em}}}{\footnotesize\textbf{ResNet-12 (scratch)}} & \multirow{2}{*}[-2.4pt]{\footnotesize\textbf{Method}} & \multicolumn{2}{c@{\hspace{1.2em}}}{\footnotesize\textbf{ResNet-18 (IN-1K)}} \\
\cmidrule(lr{11.4pt}){2-3} \cmidrule(lr{11.4pt}){5-6}
 & \makebox[\hdw][c]{\footnotesize Acc.} & \makebox[\hdw][c]{\footnotesize AA.} &  & \makebox[29.5pt][c]{\footnotesize Acc.} & \makebox[29.5pt][c]{\footnotesize AA.} \\
\midrule
PFR~\cite{pan2024pfr} & 32.42 & 40.11 & FACT~\cite{zhou2022FACT} & 69.52 & 78.52 \\
\rowrule
MEL~\cite{zhao2025mel} & 54.87 & \underline{68.49} & SAVC~\cite{song2023SAVC} & \underline{71.72} & \underline{81.29} \\
\rowrule
LGF~\cite{yin2026lgf} & \underline{56.92} & 68.31 & ADBS~\cite{li2025adbs} & 69.48 & 79.03 \\
\oursrule{6}
\rowcolor{TableHighlight}[\tabcolsep][1.2em] \rule{0pt}{12pt}\raisebox{-3pt}{\shortstack[l]{STAND\\(ours)}} & \makebox[\hdw][c]{\textbf{68.39}} & \gllapw{68.39}{+11.5}\makebox[\hdw][c]{\textbf{77.99}} & \gllapg{77.99}{+9.5}{1.2em+8pt}\rule{0pt}{12pt}\raisebox{-3pt}{\shortstack[l]{STAND\\(ours)}} & \makebox[29.5pt][c]{\textbf{76.97}} & \setlength{\hdw}{28.5pt}\gllapw{76.97}{+5.3}\makebox[\hdw][c]{\textbf{84.71}\rlap{\,\textcolor{fgGain}{\gainfmt{+3.4}}}} \\
\bottomrule
\end{tabular}}
\ifdim\wd\fgtabbox>\linewidth
  \resizebox{\linewidth}{!}{\usebox{\fgtabbox}}\else
  \makebox[\wd\fgtabbox][l]{\rlap{\usebox{\fgtabbox}}\hspace{114.6pt}\vrule height \ht\fgtabbox depth \dp\fgtabbox width .3pt}\fi
\end{table}

\noindent\textbf{Input strategy.}
Fig.~\ref{fig:design_analysis}(a) compares our composite images, dubbed Resize \& paste, with alternative input strategies while keeping our distillation unchanged. Compared with two augmented views of the original image, black patches, noise patches, and MixUp~\cite{zhang2018mixup}, also used by FunMatch~\cite{beyer2022knowledge}, remain comparable in Acc.\ and AA.\ while lowering Inc. CutMix~\cite{yun2019cutmix}, which pastes a cropped region of another image, is slightly higher in all four metrics. In contrast, Resize \& paste preserves the entire pasted image, allowing the teacher to compare its content with references drawn from whole images. Over the original views, this strategy raises Inc.\ from 52.2\% to 56.9\% and Base.\ from 73.2\% to 78.0\%, achieving the highest value in every metric. These results suggest that combining images alone is insufficient, and that preserving the content of the pasted image provides more effective inputs for relation and difference distillation.

\noindent\textbf{Projection placement.}
Fig.~\ref{fig:design_analysis}(b) examines the roles of the two projectors. The teacher projector $\gT$, trained with the classifier weights, adapts the VFM features to the visually similar base classes, while the student projector $h_\varphi$ maps the student features into the teacher's output space, so the two models are compared in one space. Compared with using neither, each projector alone improves all four metrics, raising Inc.\ from 47.2\% to 51.8\% with $\gT$ and to 55.1\% with $h_\varphi$. Using both achieves the best results, showing that adapting the teacher to the base classes and mapping the student into its space are complementary.

\begin{figure}[t]
\centering
\includegraphics[width=\linewidth]{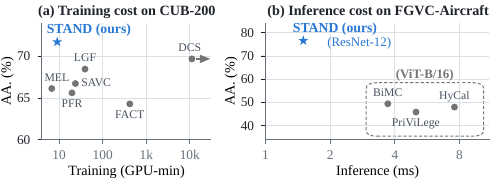}
\par\vspace{-1pt}
\caption{\textbf{Training and inference cost.} (a) uses ResNet-18~\cite{he2016resnet} (IN-1K) and counts all sessions, and the arrow marks a lower bound.}
\label{fig:cost}
\end{figure}

\noindent\textbf{Efficiency.}
Fig.~\ref{fig:cost}(a) compares our training cost on CUB-200 with recent methods~\cite{pan2024pfr,zhao2025mel,yin2026lgf,wu2025dcs} and methods that strengthen the base representation through augmentation~\cite{zhou2022FACT,song2023SAVC}. Despite using a VFM during training, ours with ResNet-18 (IN-1K) achieves the highest AA.\ in about 9 GPU minutes, including teacher preparation, close to the fastest method, MEL. At inference, methods with the same backbone run at a similar speed, so Fig.~\ref{fig:cost}(b) compares ours with methods using large pretrained backbones~\cite{park2024PriViLege,chen2025BiMC,lee2026hycal}. Ours keeps only ResNet-12, since all other modules are discarded after the base session. It runs at about 1.5 ms, more than twice as fast as these models at 3.7 to 7.6 ms, and achieves 76.60\% AA., whereas they remain below 50\% AA.\ though pretrained. These results show the practicality of STAND. Training is fast, inference is efficient, and performance is strong.

\section{Conclusion}
\label{sec:main_conclusion}

We presented STAND, a systematic design that distills VFM knowledge into a small student for FG-FSCIL. Our pilot study showed that distilled students lag behind their VFM teachers, particularly on new classes, and that changing the structure, inputs, or objectives of distillation individually barely narrows this gap. Therefore, STAND shapes them together through a fixed VFM classifier, composite images, and relation and difference distillation. Length-calibrated prototypes further restore the reduced similarity of few-shot prototypes, letting new classes compete with base classes. As a result, STAND achieves state-of-the-art performance on FG-FSCIL datasets, with every component contributing, and demonstrates flexibility across student backbones, generality on broader datasets, and efficiency. These results show that how VFM knowledge is distilled decides whether the student can stand on its own on classes it has never seen. Limitations and future directions are discussed in Supp.~Mat.

\clearpage
\section*{LLM Usage Statement}
Claude Code and Codex were used to improve the readability of the text and to help with code implementation, debugging, and figure layout.

\clearpage
\setcounter{table}{0}
\renewcommand{\thetable}{S\arabic{table}}
\setcounter{figure}{0}
\renewcommand{\thefigure}{S\arabic{figure}}
\appendix
\makeatletter
\@ifundefined{theHsection}{}{\renewcommand{\theHsection}{\Alph{section}}}
\@ifundefined{theHtable}{}{\renewcommand{\theHtable}{appendix.\arabic{table}}}
\@ifundefined{theHfigure}{}{\renewcommand{\theHfigure}{appendix.\arabic{figure}}}
\makeatother

\newcommand{\suppgroup}[1]{\par\vspace{3pt}\noindent\textbf{#1}\par}
\newcommand{\suppitem}[1]{\par\noindent\hangindent=1.7em\hangafter=1\hspace*{.7em}\textbullet\hspace{.4em}#1\par}
\widowpenalty=10000 \clubpenalty=10000 \newpage
\twocolumn[{\centering
\makesupptitle{Supplementary Material}
\vspace{0.3em}
\includegraphics[width=\textwidth]{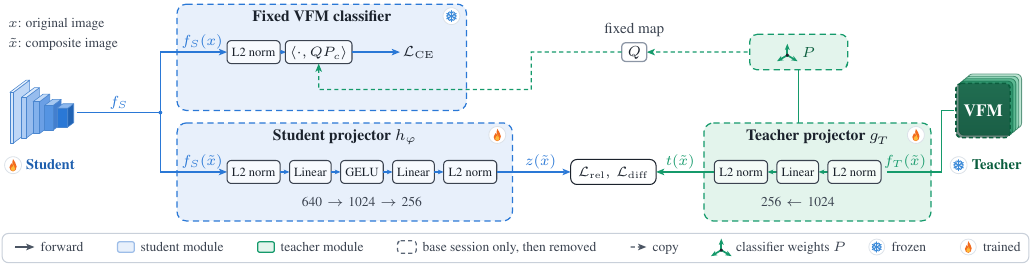}
\par\vspace{-2.5pt}
\captionof{figure}{\textbf{Projectors and fixed VFM classifier.} The teacher is DINOv3-L and the student is ResNet-12.}
\label{fig:supp:modules}
\vspace{0.5em}
}]

\noindent\colorbox{TableGroup!75!white}{\begin{minipage}{\dimexpr\columnwidth-2\fboxsep\relax}\small
This supplement supports the main paper as follows.\par
\suppgroup{\ref{sec:supp_method}. Method details}
\suppitem{Sec.~\ref{sec:supp_structure} shows the modules (Sec.~\ref{sec:main_teacher_student}).}
\suppitem{Sec.~\ref{sec:supp_difference} explains the two distillations (Secs.~\ref{sec:main_relation} and~\ref{sec:main_difference}).}
\suppitem{Sec.~\ref{sec:supp_ecp} derives and verifies the length ratio (Sec.~\ref{sec:main_calibration}).}
\suppgroup{\ref{sec:supp_analysis}. Additional analyses}
\suppitem{Sec.~\ref{sec:supp_feature_space} analyzes the feature space (Sec.~\ref{sec:main_comparison}, Tab.~\ref{tab:components}).}
\suppitem{Sec.~\ref{sec:supp_composites} examines the number of inputs (Sec.~\ref{sec:main_teacher_student}, Fig.~\ref{fig:pilot}).}
\suppitem{Sec.~\ref{sec:supp_efficiency} states the cost measurement (Fig.~\ref{fig:cost}).}
\suppitem{Sec.~\ref{sec:supp_teacher} compares teachers (Sec.~\ref{sec:main_implementation}).}
\suppgroup{\ref{sec:supp_implementation}. Experimental details}
\suppitem{Sec.~\ref{sec:supp_settings} gives the training recipes (Sec.~\ref{sec:main_implementation}).}
\suppitem{Sec.~\ref{sec:supp_baselines} describes the baselines (Fig.~\ref{fig:pilot}).}
\suppitem{Sec.~\ref{sec:supp_pretraining} lists the pretrained models (Figs.~\ref{fig:pilot} and~\ref{fig:backbones}).}
\suppitem{Sec.~\ref{sec:supp_benchmarks} describes the datasets (Sec.~\ref{sec:main_implementation}, Tab.~\ref{tab:emds7}).}
\suppgroup{\ref{sec:supp_limitations}. Limitations and future directions}
\suppgroup{\ref{sec:supp_comparison}. Detailed results\normalfont{} (Figs.~\ref{fig:teaser} and~\ref{fig:sessions}, Tab.~\ref{tab:comparison})}
\end{minipage}}\par\vspace{6pt}

\section{Method details}
\label{sec:supp_method}

\subsection{Module structure}
\label{sec:supp_structure}

Fig.~\ref{fig:supp:modules} details the modules of Sec.~\ref{sec:main_teacher_student} of the main paper.

\noindent\textbf{Fixed VFM classifier.}
The fixed VFM classifier holds the teacher's classifier weights $P=\{P_c\}$, one vector per base class $c$. These vectors lie in the 256-dimensional output space of the teacher projector $g_T$ ($d_T=256$), whereas the ResNet-12 student produces 640-dimensional features ($d_S=640$). To match the two dimensions, we use a fixed map $Q\in\mathbb{R}^{640\times256}$ to copy each $P_c$ as the vector $QP_c$. This copy does not change the similarity between any two class vectors. Since the columns of $Q$ are orthonormal, $Q^\top Q$ is the identity matrix, so $(QP_c)^\top QP_d=P_c^\top Q^\top QP_d=P_c^\top P_d$ for base classes $c$ and $d$. For an original image $x$, the classifier normalizes the student feature $f_S(x)$ and takes its similarity to each copied vector, $\langle\cdot,QP_c\rangle$, as the score of class $c$. Since the copied vectors stay frozen, the cross-entropy loss $\mathcal{L}_{\mathrm{CE}}$ on these scores trains only the student feature extractor $f_S$. Hence, the student learns to distinguish the base classes as the teacher does, with every class similarity of the teacher intact.

\noindent\textbf{Projectors.}
Unlike the classifier, the two projectors take a composite image $\tilde{x}$. The teacher projector $\gT$ is trained with the classifier weights on the base classes, so it adapts the 1024-dimensional feature $f_T(\tilde{x})$ of DINOv3-L to these classes and maps it to 256 dimensions. The student projector $h_\varphi$ maps the student feature $f_S(\tilde{x})$ into the same 256-dimensional space, so the two models are compared in one space. Each projector first normalizes its input, and its last normalization outputs $t(\tilde{x})$ for the teacher and $z(\tilde{x})$ for the student. The two distillation losses, $\mathcal{L}_{\mathrm{rel}}$ and $\mathcal{L}_{\mathrm{diff}}$, are computed between these two representations (Sec.~\ref{sec:supp_difference}).

\begin{figure*}[t]
\centering
\includegraphics[width=\textwidth]{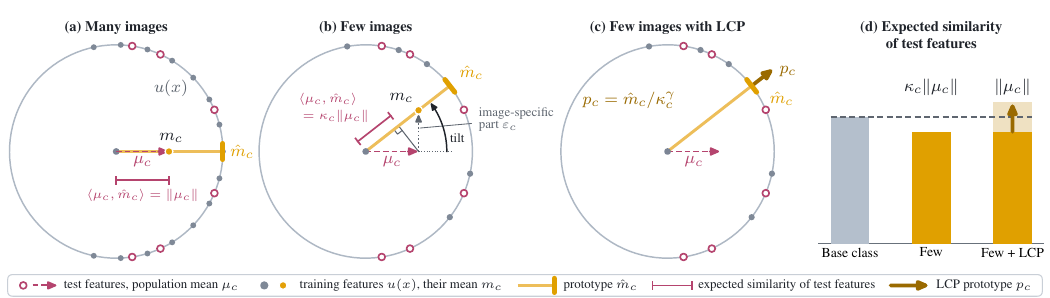}
\caption{\textbf{Prototype normalization and length calibration.} $u(x)$: unit-length feature of an image; $\mu_c$: population mean of class $c$; $m_c$: mean of its training features; $\hat{m}_c$: prototype, $m_c$ rescaled to unit length; $\kappa_c$: length ratio; $p_c$: length-calibrated prototype. The panels show the training and test features of one class, with the image-specific part $\varepsilon_c$ of $m_c$ drawn orthogonal to $\mu_c$.}
\label{fig:supp_lcp_geometry}
\end{figure*}

\subsection{Relation and difference distillation}
\label{sec:supp_difference}

We refer to the softmax distribution of the similarities between a representation and the references $b_j$ as a response.

\noindent\textbf{Relations to references.}
Sec.~\ref{sec:main_relation} of the main paper states that, by matching the responses, the student learns to compare each composite image with the references as the teacher does. Here, we make explicit what this comparison is. Consider two references $b_j,b_l\in\mathcal{R}_k$ and the representation $v$ of $\tilde{x}_k$ from either model. In the ratio of their probabilities in Eq.~\ref{eq:main_relation_distribution}, the shared denominator cancels, and taking the logarithm removes the exponentials, giving
\begin{equation}
\label{eq:supp_relation_ratio}
\tau\log\frac{q_j(v;\mathcal{R}_k)}{q_l(v;\mathcal{R}_k)}
=\langle v,b_j\rangle-\langle v,b_l\rangle.
\end{equation}
The left side depends only on the response, and the right side is how much more similar the composite image is to $b_j$ than to $b_l$. Hence, when the student's response equals the teacher's, the student has the teacher's similarity gap for every pair of references. Thus, it learns these gaps, rather than each similarity itself.

\noindent\textbf{Changes between composite images.}
For difference distillation, Sec.~\ref{sec:main_difference} states that the change in the similarity of a reference corresponds to the relative change in its probability, that is, to the factor by which its probability changes between the two composite images. We derive this correspondence here. Let $\Delta t=t(\tilde{x}_1)-t(\tilde{x}_2)$ be the difference between the teacher's representations of the two composite images. For each reference $b_j\in\mathcal{B}$, the inner product $\langle\Delta t,b_j\rangle=\langle t(\tilde{x}_1),b_j\rangle-\langle t(\tilde{x}_2),b_j\rangle$ is exactly how much more similar $b_j$ is to $\tilde{x}_1$ than to $\tilde{x}_2$. Applying softmax expresses these changes through ratios of the responses to the two composite images. To show this, we abbreviate the probability $q_j(t(\tilde{x}_k);\mathcal{R}_k)$ of reference $b_j$ in the teacher's response to $\tilde{x}_k$ as $q^T_{kj}$, and denote the denominator of this response in Eq.~\ref{eq:main_relation_distribution} by $Z_k$. First, the exponential in this equation turns the difference into a ratio, and each exponential equals $Z_k\,q^T_{kj}$ by this definition:
\begin{equation}
\label{eq:supp_difference_exponential}
\exp(\langle\Delta t,b_j\rangle/\tau)=
\frac{\exp(\langle t(\tilde{x}_1),b_j\rangle/\tau)}
{\exp(\langle t(\tilde{x}_2),b_j\rangle/\tau)}=
\frac{Z_1\,q^T_{1j}}{Z_2\,q^T_{2j}}.
\end{equation}
Next, the denominator of Eq.~\ref{eq:main_relation_distribution} normalizes this ratio over $\mathcal{B}$, and the constant $Z_1/Z_2$, which is the same for every reference, cancels:
\begin{equation}
\label{eq:supp_difference_ratio}
q_j(\Delta t;\mathcal{B})=
\frac{q^T_{1j}/q^T_{2j}}
{\sum_{b_l\in\mathcal{B}}q^T_{1l}/q^T_{2l}}.
\end{equation}
That is, the response of $\Delta t$ is the ratio of the two individual responses, normalized over the references, so it gives the most weight to the references whose probability rises by the largest factor from $\tilde{x}_2$ to $\tilde{x}_1$, regardless of how small this probability is. The response of $-\Delta t$ uses the inverse ratios, and the same identity holds for the student with $z$ in place of $t$. Hence, matching the difference responses of the two models matches how strongly the probability of each reference changes between the two composite images. Relation distillation already supervises the changes of references with large probabilities, but its KL divergence gives little weight to references whose probabilities are small in both responses, even when they change by a large factor. Difference distillation complements it on these references.

\subsection{Length-calibrated prototypes}
\label{sec:supp_ecp}

Sec.~\ref{sec:main_calibration} of the main paper states that a prototype built from a few images is less similar to test images of its class than one built from many.

\subsubsection{Why a few images lower the similarity}
\label{sec:supp_lcp_why}

To explain this drop, we assume that the unit-length features $u(x)$ of the training and test images of class $c$ are drawn independently from the same distribution, whose mean is the population mean $\mu_c$. Thus, averaging many training features or many test features gives a mean close to $\mu_c$.

Fig.~\ref{fig:supp_lcp_geometry} shows the training features of a class as filled dots on the unit circle and its test features as open dots. Their common mean $\mu_c$ lies inside the circle, so $\lVert\mu_c\rVert<1$, because unit vectors pointing in different directions partially cancel when averaged. We build the prototype $\hat{m}_c=m_c/\lVert m_c\rVert$ by rescaling the training mean $m_c$ to unit length, and measure the similarity of a test feature $u(x)$ to it by the inner product $\langle u(x),\hat{m}_c\rangle$. By the assumption, the test features average to $\mu_c$, so averaging this similarity over them gives the expected similarity $\langle\mu_c,\hat{m}_c\rangle$. Since $\hat{m}_c$ has unit length, this value is the length of the projection of $\mu_c$ onto $\hat{m}_c$.

In Fig.~\ref{fig:supp_lcp_geometry}(a), the class has many training images, as a base class does. Their image-specific variations cancel out, so the training mean $m_c$ nearly coincides with the population mean $\mu_c$, and the prototype $\hat{m}_c$ points in the direction of $\mu_c$. Therefore, the expected similarity reaches $\langle\mu_c,\hat{m}_c\rangle=\lVert\mu_c\rVert$. In Fig.~\ref{fig:supp_lcp_geometry}(b), the class has only a few training images, as a new class does. Their image-specific variations no longer cancel out, so the training mean $m_c$ keeps an image-specific part $\varepsilon_c$ in addition to $\mu_c$, that is, $m_c=\mu_c+\varepsilon_c$. On average, $\varepsilon_c$ adds nothing along $\mu_c$ (drawn as a right angle in the figure), so the inner product before normalization stays at $\langle\mu_c,m_c\rangle=\langle\mu_c,\mu_c\rangle+\langle\mu_c,\varepsilon_c\rangle=\lVert\mu_c\rVert^2$. However, $\varepsilon_c$ tilts the training mean and makes it longer than $\mu_c$. Rescaling $m_c$ to unit length divides the unchanged inner product by this longer length $\lVert m_c\rVert$, which lowers the expected similarity to $\langle\mu_c,\hat{m}_c\rangle=\lVert\mu_c\rVert^2/\lVert m_c\rVert$. The ratio of this value to the many-image value $\lVert\mu_c\rVert$ in panel (a) is the fraction of similarity retained by this prototype, the length ratio
\begin{equation}
\kappa_c=\frac{\langle\mu_c,\hat{m}_c\rangle}{\lVert\mu_c\rVert}=\frac{\lVert\mu_c\rVert^2/\lVert m_c\rVert}{\lVert\mu_c\rVert}=\frac{\lVert\mu_c\rVert}{\lVert m_c\rVert}.
\label{eq:supp_fraction_retained}
\end{equation}

In Fig.~\ref{fig:supp_lcp_geometry}(c), the length-calibrated prototype (LCP) is $p_c=\hat{m}_c/\kappa_c^\gamma$, the prototype $\hat{m}_c$ divided by $\kappa_c^\gamma$ in the same direction, where $\gamma$ is the LCP strength. By Eq.~\ref{eq:supp_fraction_retained}, the expected similarity to the prototype in panel (b) is $\langle\mu_c,\hat{m}_c\rangle=\kappa_c\lVert\mu_c\rVert$. Since the similarity is an inner product, dividing the prototype by $\kappa_c^\gamma$ divides the similarity by the same value, so the expected similarity to the LCP is
\begin{equation}
\label{eq:supp_lcp_similarity}
\langle\mu_c,p_c\rangle=\frac{\langle\mu_c,\hat{m}_c\rangle}{\kappa_c^\gamma}=\frac{\kappa_c\lVert\mu_c\rVert}{\kappa_c^\gamma}=\kappa_c^{1-\gamma}\lVert\mu_c\rVert.
\end{equation}
With $\gamma=0$, LCP leaves the prototype unchanged, and the expected similarity stays at $\kappa_c\lVert\mu_c\rVert$, as in panel (b). With $\gamma=1$, $\kappa_c$ cancels, and the expected similarity returns to $\lVert\mu_c\rVert$, the many-image value in panel (a). A value of $\gamma$ between zero and one restores part of the reduced similarity. In Fig.~\ref{fig:supp_lcp_geometry}(d), the \emph{Few} bar is below the \emph{Base class} bar, so the new class can lose to the base class. The \emph{Few + LCP} bar rises above it to $\lVert\mu_c\rVert$, the value in Fig.~\ref{fig:supp_lcp_geometry}(a).

\subsubsection{Estimation from training images}
\label{sec:supp_lcp_derivation}
Eq.~\ref{eq:supp_fraction_retained} requires $\lVert\mu_c\rVert$, but in practice we cannot know the population mean $\mu_c$. Instead, we estimate $\lVert\mu_c\rVert^2$ from the $n_c$ training features $u_1,\dots,u_{n_c}$ of class $c$, which average to the training mean $m_c=\frac{1}{n_c}\sum_{i=1}^{n_c}u_i$. Under the assumption of Sec.~\ref{sec:supp_lcp_why}, the features $u_i$ and $u_j$ of two distinct images are independent with mean $\mu_c$, so their expected inner product is
\begin{equation}
\label{eq:supp_pair_expectation}
\mathbb{E}\langle u_i,u_j\rangle=\langle\mathbb{E}[u_i],\mathbb{E}[u_j]\rangle=\langle\mu_c,\mu_c\rangle=\lVert\mu_c\rVert^2,
\end{equation}
regardless of $n_c$. Hence, we can estimate $\lVert\mu_c\rVert^2$ by the mean similarity $\widehat{\rho}_c$ between distinct images of class $c$ in Eq.~\ref{eq:main_lcp_ratio} of the main paper, that is, the mean inner product over all $n_c(n_c-1)$ pairs of these images,
\begin{equation}
\label{eq:supp_rho_pairs}
\widehat{\rho}_c=\frac{1}{n_c(n_c-1)}\sum_{i\neq j}\langle u_i,u_j\rangle.
\end{equation}
Eq.~\ref{eq:supp_rho_pairs} defines the mean similarity $\widehat{\rho}_c$, but evaluating it directly requires the inner products of all pairs. Instead, since we already know the training mean $m_c$, we can derive the same value from the length $\lVert m_c\rVert$ of this mean. This is possible because $m_c$ averages the same features, so its squared length expands into the inner products of all $n_c^2$ pairs, which include the pairs of Eq.~\ref{eq:supp_rho_pairs},
\begin{equation}
\label{eq:supp_mean_length}
\begin{aligned}
\lVert m_c\rVert^2&=\langle m_c,m_c\rangle=\Bigl\langle\frac{1}{n_c}\sum_{i}u_i,\frac{1}{n_c}\sum_{j}u_j\Bigr\rangle\\
&=\frac{1}{n_c^2}\sum_{i}\sum_{j}\langle u_i,u_j\rangle\\
&=\frac{1}{n_c^2}\Bigl(\underbrace{\sum_{i}\langle u_i,u_i\rangle}_{n_c}+\underbrace{\sum_{i\neq j}\langle u_i,u_j\rangle}_{n_c(n_c-1)\widehat{\rho}_c}\Bigr)\\
&=\frac{n_c}{n_c^2}+\frac{n_c(n_c-1)}{n_c^2}\,\widehat{\rho}_c\\
&=\frac{1}{n_c}+\Bigl(1-\frac{1}{n_c}\Bigr)\widehat{\rho}_c.
\end{aligned}
\end{equation}
Here, each of the $n_c$ pairs with $i=j$ gives $\langle u_i,u_i\rangle=1$, since the features have unit length, and the remaining pairs sum to $n_c(n_c-1)\widehat{\rho}_c$ by Eq.~\ref{eq:supp_rho_pairs}. Multiplying Eq.~\ref{eq:supp_mean_length} by $n_c$ gives $n_c\lVert m_c\rVert^2=1+(n_c-1)\widehat{\rho}_c$, so for $n_c>1$,
\begin{equation}
\label{eq:supp_rho_length}
\widehat{\rho}_c=\frac{n_c\lVert m_c\rVert^2-1}{n_c-1},
\end{equation}
\begin{table}[t]
\caption{\textbf{New-class prototypes built from different numbers of images.} Inner: inner product of test features with the mean $m_c$; Length: length of $m_c$; Similarity: similarity of test features to the prototype $\hat{m}_c$, which equals Inner divided by Length.}
\label{tab:supp:lcp_image_count}
\centering
\setlength{\tabcolsep}{3pt}
\renewcommand{\arraystretch}{1.2}
\scriptsize
\settowidth{\hdw}{\footnotesize Similarity}
\sbox{\fgtabbox}{\begin{tabular}{ccccccc}
\toprule
\multirow{2}{*}[-2.4pt]{\footnotesize\textbf{Images}} & \multicolumn{3}{c}{\footnotesize\textbf{FGVC-Aircraft}} & \multicolumn{3}{c}{\footnotesize\textbf{Stanford Cars}} \\
\cmidrule(lr){2-4} \cmidrule(lr){5-7}
 & \makebox[\hdw][c]{\footnotesize Inner} & \makebox[\hdw][c]{\footnotesize Length} & \makebox[\hdw][c]{\footnotesize Similarity} & \makebox[\hdw][c]{\footnotesize Inner} & \makebox[\hdw][c]{\footnotesize Length} & \makebox[\hdw][c]{\footnotesize Similarity} \\
\midrule
1 & 0.40 & 1.00 & 0.40 & 0.36 & 1.00 & 0.36 \\
\rowrule
2 & 0.40 & 0.84 & 0.48 & 0.36 & 0.82 & 0.44 \\
\rowrule
3 & 0.40 & 0.77 & 0.52 & 0.36 & 0.76 & 0.47 \\
\rowrule
5 & 0.40 & 0.72 & 0.55 & 0.36 & 0.70 & 0.51 \\
\rowrule
10 & 0.40 & 0.68 & 0.59 & 0.36 & 0.65 & 0.55 \\
\rowrule
20 & 0.40 & 0.65 & 0.61 & 0.36 & 0.63 & 0.57 \\
\bottomrule
\end{tabular}}
\ifdim\wd\fgtabbox>\linewidth
  \resizebox{\linewidth}{!}{\usebox{\fgtabbox}}\else
  \usebox{\fgtabbox}\fi
\end{table}
\unskip\begin{figure*}[t]
\centering
\includegraphics[width=\linewidth]{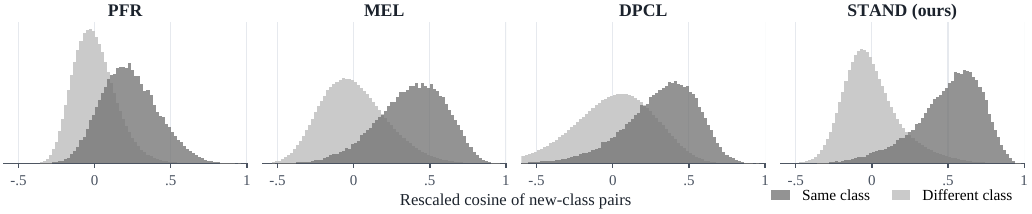}
\par\vspace{-3pt}
\caption{\textbf{New-class feature similarities.} Cosine similarity of all new-class test image pairs, computed with each method's feature extractor on FGVC-Aircraft and rescaled per method so that the different-class mean is 0 and identical pairs are 1.}
\label{fig:supp:feature_separation}
\vspace{11pt}\end{figure*}which needs only the length of the training mean. Then, replacing $\lVert\mu_c\rVert$ in Eq.~\ref{eq:supp_fraction_retained} by $\sqrt{\widehat{\rho}_c}$ gives the length ratio $\kappa_c=\sqrt{\widehat{\rho}_c}/\lVert m_c\rVert$ in Eq.~\ref{eq:main_lcp_ratio} of the main paper. Moreover, Eq.~\ref{eq:supp_mean_length} can be rewritten as $\lVert m_c\rVert^2=\widehat{\rho}_c+(1-\widehat{\rho}_c)/n_c$, so the squared length of the training mean exceeds $\widehat{\rho}_c$ by $(1-\widehat{\rho}_c)/n_c$, and the length ratio is
\begin{equation}
\label{eq:supp_kappa_form}
\kappa_c=\frac{\sqrt{\widehat{\rho}_c}}{\lVert m_c\rVert}=\sqrt{\frac{\widehat{\rho}_c}{\widehat{\rho}_c+(1-\widehat{\rho}_c)/n_c}}.
\end{equation}
As the number of images $n_c$ grows, the term $(1-\widehat{\rho}_c)/n_c$ vanishes, so $\kappa_c$ approaches one. In contrast, $\kappa_c$ is smaller for fewer or less similar images, that is, for a smaller number of images $n_c$ or a smaller mean similarity $\widehat{\rho}_c$.

\subsubsection{Verification on real features}
\label{sec:supp_lcp_check}

Tab.~\ref{tab:supp:lcp_image_count} verifies the explanation of Sec.~\ref{sec:supp_lcp_why} on real features. It measures the inner product of test features with the mean $m_c$ as \mbox{\emph{Inner}} and the length of $m_c$ as \emph{Length}. Their ratio is the similarity of the test features to the prototype $\hat{m}_c=m_c/\lVert m_c\rVert$, reported as \emph{Similarity}. Since a new class has only five training images and its test images follow the same distribution by our assumption, we build this mean from its test images instead, for this verification only. Specifically, for each new class, we randomly pick one to twenty of its test images to build the mean $m_c$ and score its remaining test images. We repeat this with 20 random picks and report the average. As the mean is built from more images, \mbox{\emph{Inner}} stays constant, at 0.40 on FGVC-Aircraft and 0.36 on Stanford Cars, which confirms that the image-specific part $\varepsilon_c$ adds nothing along $\mu_c$ on average, as drawn in Fig.~\ref{fig:supp_lcp_geometry}(b). In contrast, \emph{Length} falls from 1.00 to 0.65 on FGVC-Aircraft, which raises \emph{Similarity} from 0.40 to 0.61. Thus, a longer mean lowers the similarity of a few-image prototype to the test features of its class.

In practice, each new class has only its five training images. The length ratio $\kappa_c$ computed from these images averages 0.86, 0.85, 0.80, and 0.72 on FGVC-Aircraft, Stanford Cars, CUB-200, and Stanford Dogs, close to the fractions of similarity that their prototypes keep, 0.87, 0.86, 0.80, and 0.72, which we measure by using the mean of the test features in place of $\mu_c$ in Eq.~\ref{eq:supp_fraction_retained}. Thus, $\kappa_c$ estimates this reduction closely from the training images alone. For base classes, which have 30 to 100 training images, $\kappa_c$ averages 0.99 on all four datasets. Hence, LCP mainly lengthens new-class prototypes and leaves base-class prototypes nearly unchanged.

\section{Additional analyses}
\label{sec:supp_analysis}

\subsection{Feature-space analysis}
\label{sec:supp_feature_space}

\begin{figure}[t]
\centering
\includegraphics[width=\linewidth]{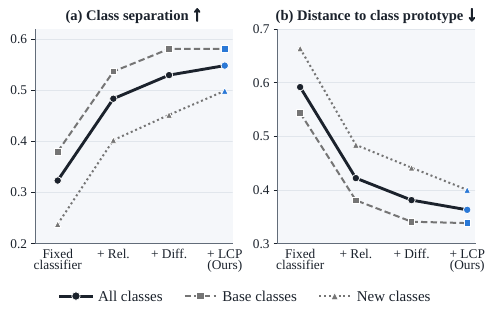}
\par\vspace{-5pt}
\caption{\textbf{Class separation and distance to class prototype on FGVC-Aircraft.} Class separation: how much closer test images lie to their class prototype than to all prototypes; Distance: how far they lie from their class prototype.}
\label{fig:supp:prototype_geometry}
\end{figure}

\begin{figure*}[t]
\centering
\includegraphics[width=\linewidth]{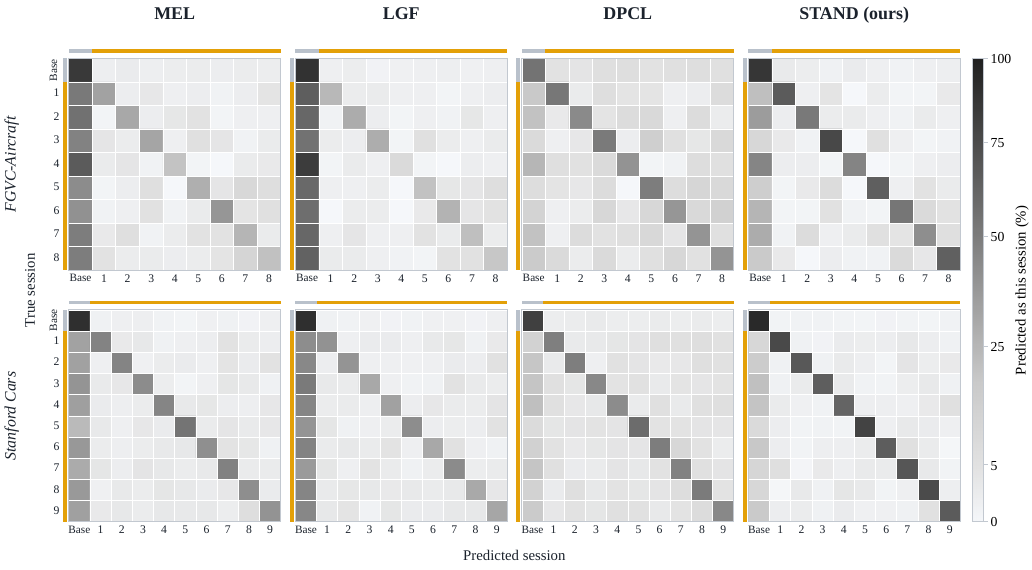}
\par\vspace{-5pt}
\caption{\textbf{Final-session confusion matrices.} Classes are grouped by the session that introduced them, and each row gives how the test images of one session are predicted, summing to 100\%. Gray and orange bands mark base and new classes.}
\label{fig:supp:confusion}
\end{figure*}

Following the accuracy comparison in Sec.~\ref{sec:main_comparison} of the main paper, we analyze the features and predictions behind it.

\noindent\textbf{Feature similarity.}
Since the base representation is not updated for new classes, it must already capture the subtle details that separate them. Fig.~\ref{fig:supp:feature_separation} measures how well each method separates new classes. It plots the cosine similarities of same-class and different-class pairs of new-class test images, computed with each method's feature extractor. After rescaling, different-class pairs center at zero, so capturing these details means moving same-class pairs toward one. With PFR~\cite{pan2024pfr}, MEL~\cite{zhao2025mel}, and DPCL~\cite{mao2026dpcl}, same-class pairs stay closer to zero, so many of them are no more similar than different-class pairs. In contrast, with STAND (ours), same-class pairs lie the closest to one, while different-class pairs stay near zero. Thus, ours draws images of the same new class together and keeps different new classes apart, although fine-grained classes differ only in subtle details. This separation aligns with Sec.~\ref{sec:main_comparison}, where the advantage of ours over the other methods grows as new classes arrive.

\noindent\textbf{Class separation and distance.}
To see which components bring this separation, Fig.~\ref{fig:supp:prototype_geometry} adds them one at a time and measures two quantities. Class separation (higher is better) is how much closer a test image lies to its class prototype than to all prototypes on average. The distance (lower is better) is how far the image lies from its class prototype. As in Tab.~\ref{tab:components} of the main paper, relation distillation brings the largest step, raising class separation and reducing the distance for both base and new classes. Difference distillation adds a further step in both panels. Finally, LCP lengthens new-class prototypes more than base-class prototypes. In this last step, the new-class lines keep improving in both panels, whereas the base-class lines plateau. Overall, each component raises class separation and reduces the distance over all classes, which matches Tab.~\ref{tab:components}, where each component improves Acc.\ and AA.\ on both datasets.

\noindent\textbf{Final-session predictions.}
Fig.~\ref{fig:supp:confusion} shows the predictions in the final session. For MEL and LGF~\cite{yin2026lgf}, the Base column is dark, meaning that many new-class images are predicted as base classes. DPCL lightens this column, but its top-left cell is relatively light, since it predicts more base-class images as new classes than the others. In contrast, ours keeps this cell dark, like MEL and LGF, and has the darkest diagonal cell in every incremental session. Altogether, ours keeps classes apart in its features and predicts the images of both base and new classes in the correct session.

\subsection{Composite images and augmentation}
\label{sec:supp_composites}

Sec.~\ref{sec:main_teacher_student} of the main paper builds two composite images per original image as the inputs for distillation. Since composite images also give the student more inputs per original image, we examine whether our gain comes from the number of inputs alone. For this purpose, Tab.~\ref{tab:supp:kd_augmentation} compares ours with KD~\cite{hinton2015distilling} trained with more inputs.

\begin{table}[t]
\caption{\textbf{Comparison of the number of inputs on FGVC-Aircraft.} Views: the number of augmented copies per original image.}
\label{tab:supp:kd_augmentation}
\centering
\scriptsize
\setlength{\tabcolsep}{1pt}
\renewcommand{\arraystretch}{1.2}
\def\fgaugrows{\toprule
\raisebox{4.6pt}{\footnotesize\textbf{Condition}} & \footnotesize\textbf{\shortstack{Training\\(hours)}} & \raisebox{4.6pt}{\footnotesize\textbf{Base.}} & \raisebox{4.6pt}{\footnotesize\textbf{Inc.}} & \raisebox{4.6pt}{\footnotesize\textbf{Acc.}} & \raisebox{4.6pt}{\footnotesize\textbf{AA.}} \\
\midrule
KD, 1 view & 0.63 & 70.03 & 44.70 & 59.90 & 67.79 \\
\rowrule
KD, 2 views & 1.11 & 72.90 & 51.10 & 64.18 & 71.73 \\
\rowrule
KD, 4 views & 2.18 & 74.17 & 53.90 & 66.06 & 73.31 \\
\rowrule
KD, 8 views & 4.29 & 75.40 & 53.85 & 66.78 & 74.52 \\
\oursrule{6}
\rowcolor{TableHighlight} STAND (ours) & 1.52 & \textbf{78.03} & \textbf{56.85} & \textbf{69.56} & \textbf{77.02} \\
\bottomrule
}
\sbox{\fgtabbox}{\begin{tabular}{lccccc}\fgaugrows\end{tabular}}
\addtolength{\tabcolsep}{\dimexpr(\linewidth-\wd\fgtabbox)/12\relax}
\begin{tabular}{lccccc}\fgaugrows\end{tabular}
\end{table}

\begin{figure*}[t]
\centering
\includegraphics[width=\textwidth]{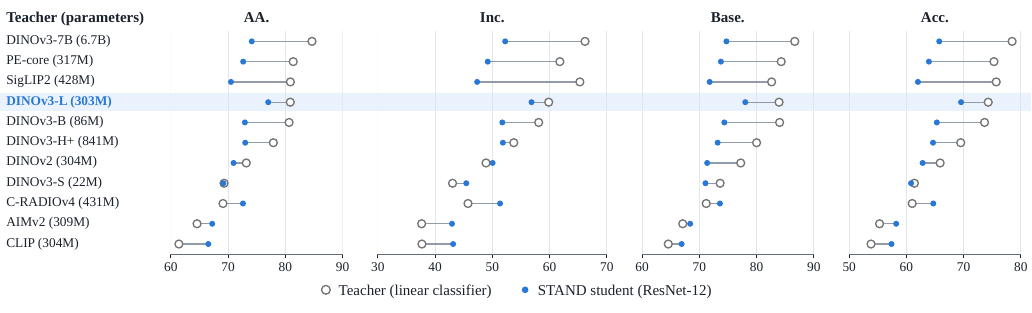}
\par\vspace{-4pt}
\caption{\textbf{Teacher comparison on FGVC-Aircraft.} Each row shows the accuracy (\%) of a teacher with its linear classifier and after transfer to its student.}
\label{fig:supp:teachers}
\end{figure*}

Specifically, we train KD with one, two, four, and eight augmented views per original image. A view is a copy of the original image with its own random crop, flip, and AutoAugment~\cite{cubuk2019autoaugment}. More views raise all four metrics of KD, so the inputs matter for distillation. However, its gain on new classes has a limit. From four to eight views, the training time doubles and Base.\ still rises, but Inc.\ stays at about 53.9\%. In contrast, ours reaches 78.03\% Base.\ and 56.85\% Inc.\ in 1.52 training hours, above the 75.40\% and 53.85\% that KD reaches with eight views in 4.29 hours. Thus, more inputs help distillation, but increasing their number alone reaches a limit on Inc., whereas ours exceeds this limit in less training time.

\subsection{Cost details}
\label{sec:supp_efficiency}

The training time of ours in Fig.~\ref{fig:cost}(a) of the main paper counts four steps: teacher feature extraction, training of the teacher projector and classifier weights, student training, and post-processing. Student training takes 8.51 of the 9.03 GPU minutes, including the teacher's forward passes. Pretraining on public data is not counted for any method. The DCS~\cite{wu2025dcs} value is a lower bound. It counts its image generation and classifier training but leaves out its base-session fine-tuning and model loading. Fig.~\ref{fig:cost}(b) measures the inference time for one image, from the preprocessed image to the predicted class. Each method runs its whole network and classifier at its training input size, $84\times84$ for ours with ResNet-12 and $224\times224$ for the compared methods with ViT-B/16.

\subsection{Teacher selection}
\label{sec:supp_teacher}

Sec.~\ref{sec:main_implementation} of the main paper uses DINOv3-L~\cite{simeoni2025dinov3} as the default teacher, and Fig.~\ref{fig:supp:teachers} compares it with ten other VFMs. With each of the eleven teachers, our student exceeds every method in Tab.~\ref{tab:comparison} of the main paper in AA., so STAND is not tied to one VFM. However, a more accurate or larger VFM is not necessarily a better teacher. DINOv3-7B is the most accurate teacher in all four metrics and 22 times larger than DINOv3-L, but its student is below the DINOv3-L student in every metric. Likewise, PE-core~\cite{bolya2025perception} is more accurate than DINOv3-L and DINOv3-H+ is larger, yet both yield weaker students. This agrees with prior observations that a more accurate teacher~\cite{cho2019efficacy} or a larger one~\cite{cho2019efficacy,mirzadeh2020improved} does not necessarily give a better student. Therefore, we choose the VFM by how well its knowledge reaches the student. DINOv3-L yields the best student in all four metrics, so it is our default.

\section{Experimental details}
\label{sec:supp_implementation}

\subsection{Implementation details}
\label{sec:supp_settings}

\noindent\textbf{Training recipes.}
We first train the teacher projector $\gT$ and the classifier weights $P$ on frozen VFM features of base images for 60 epochs with AdamW~\cite{loshchilov2017AdamW}, a learning rate of $10^{-3}$, a weight decay of $5\times10^{-4}$, and a batch size of 128. During distillation, both models receive the same original and composite images, resized for the teacher to the input size it requires. The student uses the distillation weights $\lambda_r=\lambda_d=1$ and the LCP strength $\gamma=0.5$ on every dataset, while its optimizer depends on the backbone. ResNet-12 (scratch) uses SGD with Nesterov momentum 0.9, a weight decay of $5\times10^{-4}$, and a learning rate of 0.3 with a cosine schedule. ResNet-18 (scratch) uses SGD with momentum 0.9, a weight decay of $10^{-4}$, and a learning rate of 0.1 after a five-epoch warm-up. Pretrained students use a smaller learning rate for their pretrained backbone than for the layers added on top of it, with a cosine schedule. Specifically, the ResNet and MobileNetV2 students use the SGD of ResNet-12 with rates of 0.002 and 0.02, and the ViT-B/16 students use AdamW with a weight decay of 0.05 and rates of $10^{-4}$ and $10^{-3}$. Students trained from scratch run for 400 epochs, and pretrained ones for 30 to 60 epochs.

\begin{figure*}[t]
\centering
\includegraphics[width=\textwidth]{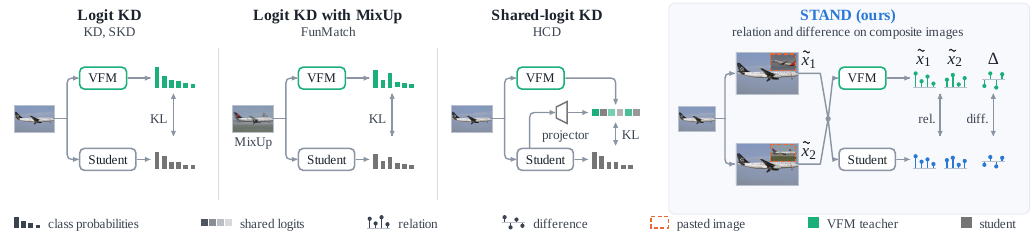}
\par\vspace{-3pt}
\caption{\textbf{What each distillation family matches.} The first three panels show the compared distillation methods and the last shows ours. Vertical arrows mark what the student matches to the teacher, and rel.\ and diff.\ denote relation and difference distillation.}
\label{fig:supp:distillation_comparison}
\end{figure*}

\begin{figure*}[t]
\centering
\includegraphics[width=\linewidth]{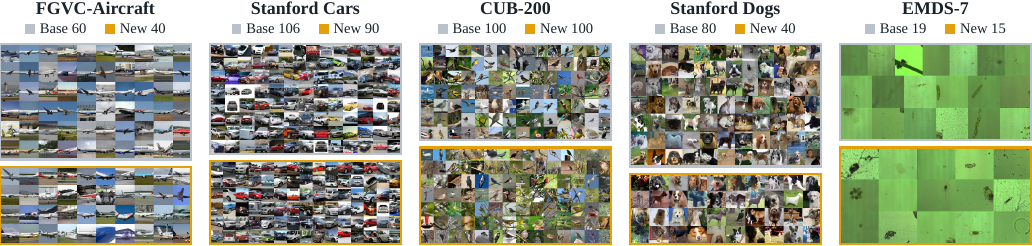}
\par\vspace{-1.5pt}
\caption{\textbf{The four fine-grained datasets and EMDS-7.} Each tile is one training image of one class, in class order, with base classes framed in gray and new classes framed in orange.}
\label{fig:supp:benchmarks}
\end{figure*}

\noindent\textbf{Inputs.}
Following PFR~\cite{pan2024pfr}, students trained from scratch take $84\times84$ inputs with AutoAugment~\cite{cubuk2019autoaugment} on FGVC-Aircraft, Stanford Cars, and Stanford Dogs, and $224\times224$ inputs without AutoAugment on CUB-200. These students use the $84\times84$ setting on CIFAR-100 and mini\-ImageNet and the $224\times224$ setting on EMDS-7. Pretrained students take $224\times224$ inputs without AutoAugment on every dataset.

\noindent\textbf{Composite images.}
The composite images in Sec.~\ref{sec:main_teacher_student} of the main paper use the Resize \& paste operation of ResizeMix~\cite{qin2020resizemix}. We resize the whole pasted image by a ratio drawn uniformly from $[0.25,0.75]$ and paste it at a random location in the background image.

\subsection{Distillation baselines}
\label{sec:supp_baselines}

Fig.~\ref{fig:pilot}(b) of the main paper compares four distillation methods, KD, SKD, FunMatch, and HCD, which differ in what the student matches, as Fig.~\ref{fig:supp:distillation_comparison} shows for each family. Logit KD matches class probabilities: KD~\cite{hinton2015distilling} does only this, SKD~\cite{jeong2026skd} also matches the cosine similarities between the logits of images in a batch, and FunMatch~\cite{beyer2022knowledge} matches class probabilities on MixUp~\cite{zhang2018mixup} images. For shared-logit KD, HCD~\cite{xu2026hcd} joins the student's intermediate features with the teacher's features into shared logits that the student's class probabilities match. In contrast, STAND (ours) feeds both models composite images and matches how the teacher relates each of them to other images and how this relation changes between two of them.

\subsection{Pretrained models}
\label{sec:supp_pretraining}

\noindent\textbf{Teachers.}
The default teacher uses DINOv3-L\urlnote{hf.co/timm/vit_large_patch16_dinov3.lvd1689m}{huggingface.co/timm/vit_large_patch16_dinov3.lvd1689m}~\cite{simeoni2025dinov3}. The other teachers in Fig.~\ref{fig:pilot} of the main paper and Fig.~\ref{fig:supp:teachers} are DINOv3-S\urlnote{hf.co/timm/vit_small_patch16_dinov3.lvd1689m}{huggingface.co/timm/vit_small_patch16_dinov3.lvd1689m}, DINOv3-B\urlnote{hf.co/timm/vit_base_patch16_dinov3.lvd1689m}{huggingface.co/timm/vit_base_patch16_dinov3.lvd1689m}, DINOv3-H+\urlnote{hf.co/timm/vit_huge_plus_patch16_dinov3.lvd1689m}{huggingface.co/timm/vit_huge_plus_patch16_dinov3.lvd1689m}, DINOv3-7B\urlnote{hf.co/timm/vit_7b_patch16_dinov3.lvd1689m}{huggingface.co/timm/vit_7b_patch16_dinov3.lvd1689m}, DINOv2\urlnote{hf.co/timm/vit_large_patch14_dinov2.lvd142m}{huggingface.co/timm/vit_large_patch14_dinov2.lvd142m}~\cite{oquab2023DINOv2}, PE-core\urlnote{hf.co/timm/vit_pe_core_large_patch14_336.fb}{huggingface.co/timm/vit_pe_core_large_patch14_336.fb}~\cite{bolya2025perception}, C-RADIOv4\urlnote{hf.co/nvidia/C-RADIOv4-SO400M}{huggingface.co/nvidia/C-RADIOv4-SO400M}~\cite{ranzinger2026cradio}, SigLIP2\urlnote{hf.co/timm/vit_so400m_patch14_siglip_224.v2_webli}{huggingface.co/timm/vit_so400m_patch14_siglip_224.v2_webli}~\cite{tschannen2025siglip2}, AIMv2\urlnote{hf.co/timm/aimv2_large_patch14_224.apple_pt}{huggingface.co/timm/aimv2_large_patch14_224.apple_pt}~\cite{fini2025aimv2}, and CLIP\urlnote{github.com/openai/CLIP}{github.com/openai/CLIP}~\cite{radford2021CLIP}.

\noindent\textbf{Students.}
ResNet-12 (scratch) and ResNet-18 (scratch) start from random weights. MobileNetV2~\cite{sandler2018mobilenetv2}, ResNet-18, and ResNet-50~\cite{he2016resnet} marked (IN-1K) start from torchvision\urlnote{pytorch.org/vision/stable/models.html}{pytorch.org/vision/stable/models.html} weights pretrained on ImageNet-1K~\cite{deng2009imagenet}, and ViT-B/16~\cite{dosovitskiy2021vit} marked (IN-21K) starts from timm\urlnote{hf.co/timm/vit_base_patch16_224.augreg_in21k_ft_in1k}{huggingface.co/timm/vit_base_patch16_224.augreg_in21k_ft_in1k} weights pretrained on ImageNet-21K and fine-tuned on ImageNet-1K.

\subsection{Dataset details}
\label{sec:supp_benchmarks}

Fig.~\ref{fig:supp:benchmarks} shows one training image of every class in each dataset. In the four fine-grained datasets, the classes differ from one another only in subtle details. For the sessions in Sec.~\ref{sec:main_implementation} of the main paper, we use the class and image lists of PFR~\cite{pan2024pfr} on FGVC-Aircraft, Stanford Cars, and Stanford Dogs, and those of CEC~\cite{zhang2021CEC} on CUB-200, CIFAR-100, and miniImageNet. With these lists, each base class has 50 training images on FGVC-Aircraft, 40 on Stanford Cars, 30 on CUB-200, 100 on Stanford Dogs, and 500 on CIFAR-100 and miniImageNet. For EMDS-7~\cite{yang2023emds7}, we use 19 base classes of 20 training images, followed by 5 incremental sessions in the 3-way, 3-shot setting of Zhou et al.~\cite{zhou2026emds7}, and train all methods in Tab.~\ref{tab:emds7} of the main paper on this split.

\section{Limitations and future directions}
\label{sec:supp_limitations}

Like most FSCIL methods~\cite{zhang2021CEC,zhou2022FACT}, we freeze the student after the base session, since updating it with limited images causes overfitting and forgetting. Consequently, the student cannot draw additional VFM knowledge while learning new classes. Moreover, since a larger or more accurate VFM does not always give a better student (Fig.~\ref{fig:supp:teachers}), searching for the teacher in advance can be important. To address these limitations, a promising direction for future work is to continue distillation in incremental sessions, transferring only the VFM knowledge relevant to the new classes and updating only a small part of the student. Another direction is to distill from several VFMs rather than a single one so that they complement each other.

\makeatletter\begingroup\let\lp@need\relax\@secpenalty=\@M
\section{Detailed results}\endgroup\makeatother
\label{sec:supp_comparison}
\label{sec:supp_sessions}

\begingroup
\makeatletter
\let\fgfloat\table
\let\endfgfloat\endtable
\fgfloat[!b]
\renewenvironment{table}[1][]{\par\noindent\begin{minipage}{\textwidth}\def\@captype{table}}{\end{minipage}\par}
\setbox0=\vbox{\hsize=\textwidth\linewidth=\textwidth\parskip=0pt\parindent=0pt
\begin{table}[!ht]
\caption{\textbf{Per-session accuracy on FGVC-Aircraft.} Our results are averaged over three seeds. Acc./AA.: final/average accuracy; Base./Inc.: final base/new-class accuracy (\%). $\pm$: standard deviation. \textbf{Bold}/\underline{underline}: best/second among incremental methods; -: unavailable. Joint~Learning: trained on the training images of all sessions at once.}
\label{tab:supp:sessions_aircraft}
\centering
\setlength{\tabcolsep}{4.86pt}
\renewcommand{\arraystretch}{1.2}
\scriptsize
\settowidth{\hdw}{\footnotesize\textbf{Base.}}
\sbox{\fgtabbox}{\begin{tabular}{lllcccccccccccc@{\hspace{1.3em}}}
\toprule
\multirow{2}{*}[-2.3pt]{\footnotesize\textbf{Method}} & \multirow{2}{*}[-2.3pt]{\footnotesize\textbf{Venue}} & \multirow{2}{*}[-2.3pt]{\footnotesize\textbf{Backbone}} & \multicolumn{8}{c}{\footnotesize\textbf{Sessions}} & \multicolumn{3}{c}{\footnotesize\textbf{Final session (8)}} & \multirow{2}{*}[-2.3pt]{\footnotesize\textbf{AA.}} \\
\cmidrule(lr){4-11} \cmidrule(lr){12-14}
 &  &  & \footnotesize\textbf{0} & \footnotesize\textbf{1} & \footnotesize\textbf{2} & \footnotesize\textbf{3} & \footnotesize\textbf{4} & \footnotesize\textbf{5} & \footnotesize\textbf{6} & \footnotesize\textbf{7} & \makebox[\hdw][c]{\footnotesize\textbf{Acc.}} & \makebox[\hdw][c]{\footnotesize\textbf{Base.}} & \makebox[\hdw][c]{\footnotesize\textbf{Inc.}} &  \\
\midrule
Joint Learning &  & \diagcell{ResNet-12 (scratch)} &  &  &  &  &  &  &  &  & 48.36 &  &  & \\
\midrule
\rowcolor{TableGroup}[\tabcolsep][\tabcolsep] \multicolumn{15}{l}{\textit{FSCIL methods}} \\
CEC~\cite{zhang2021CEC} & CVPR'21 & \diagcell{ResNet-12 (scratch)} & 50.61 & 43.50 & 40.32 & 37.26 & 34.93 & 33.95 & 32.15 & 30.17 & 29.57 & - & - & 36.94 \\
\rowrule
FACT~\cite{zhou2022FACT} & CVPR'22 & \diagcell{ResNet-12 (scratch)} & 55.95 & 49.19 & 45.25 & 41.53 & 38.06 & 35.58 & 34.31 & 32.64 & 31.60 & - & - & 40.46 \\
\rowrule
SAVC~\cite{song2023SAVC} & CVPR'23 & \diagcell{ResNet-18 (IN-1K)} & - & - & - & - & - & - & - & - & \underline{55.26} & 69.40 & 34.05 & \underline{64.26} \\
\rowrule
TEEN~\cite{wang2023teen} & NeurIPS'23 & \diagcell{ResNet-12 (scratch)} & 59.17 & 55.08 & 50.06 & 46.05 & 42.10 & 38.73 & 35.60 & 33.01 & 30.98 & - & - & 43.42 \\
\rowrule
PriViLege~\cite{park2024PriViLege} & CVPR'24 & \diagcell{ViT-B/16 (IN-21K)} & - & - & - & - & - & - & - & - & 38.56 & 41.77 & 33.75 & 45.87 \\
\rowrule
CLOSER~\cite{oh2024CLOSER} & ECCV'24 & \diagcell{ResNet-12 (scratch)} & 68.07 & 57.45 & 52.86 & 49.28 & 46.05 & 43.20 & 40.73 & 38.51 & 36.56 & - & - & 48.08 \\
\rowrule
ADBS~\cite{li2025adbs} & AAAI'25 & \diagcell{ResNet-12 (scratch)} & 59.03 & 53.67 & 50.60 & 46.61 & 44.28 & 41.19 & 39.36 & 36.90 & 35.42 & - & - & 45.23 \\
\rowrule
BiMC~\cite{chen2025BiMC} & CVPR'25 & \diagcell{ViT-B/16 (CLIP)} & - & - & - & - & - & - & - & - & 42.82 & 47.33 & 36.05 & 49.45 \\
\rowrule
HyCal~\cite{lee2026hycal} & CVPR'26 & \diagcell{ViT-B/16 (CLIP)} & - & - & - & - & - & - & - & - & 40.06 & 42.43 & 36.50 & 48.03 \\
\rowrule
DPCL~\cite{mao2026dpcl} & CVPRF'26 & \diagcell{ResNet-12 (scratch)} & \underline{71.50} & 63.67 & 59.66 & 57.13 & 53.23 & 52.14 & 49.84 & 48.23 & 46.12 & 49.93 & 40.40 & 55.72 \\
\addlinespace[2pt]
\rowcolor{TableGroup}[\tabcolsep][\tabcolsep] \multicolumn{15}{l}{\textit{Fine-grained FSCIL methods}} \\
PFR~\cite{pan2024pfr} & PR'24 & \diagcell{ResNet-12 (scratch)} & 59.94 & 52.80 & 47.66 & 43.32 & 39.61 & 38.16 & 36.36 & 35.22 & 34.42 & - & - & 43.05 \\
\rowrule
PE-Net~\cite{nie2025penet} & ICONIP'25 & \diagcell{ResNet-12 (scratch)} & 67.30 & 61.82 & 56.71 & 51.37 & 47.86 & 44.19 & 41.53 & 38.44 & 36.56 & - & - & 49.53 \\
\rowrule
MEL~\cite{zhao2025mel} & NeurIPS'25 & \diagcell{ResNet-12 (scratch)} & 70.93 & \underline{67.12} & \underline{63.91} & \underline{60.79} & \underline{57.75} & \underline{55.86} & \underline{54.49} & \underline{52.15} & 50.06 & 65.03 & 27.60 & 59.23 \\
\rowrule
LGF~\cite{yin2026lgf} & IJMIR'26 & \diagcell{ResNet-12 (scratch)} & 66.30 & 62.36 & 59.83 & 57.26 & 54.10 & 51.56 & 49.96 & 47.98 & 46.28 & 62.90 & 21.35 & 55.07 \\
\addlinespace[2pt]
\rowcolor{TableHighlight}[\tabcolsep][1.3em] STAND (ours) &  & \diagcell{ResNet-12 (scratch)} & \textbf{85.41} & \textbf{83.02} & \textbf{79.48} & \textbf{78.74} & \textbf{75.52} & \textbf{74.26} & \textbf{72.93} & \textbf{70.78} & \textbf{69.23} & 77.92 & 56.20 & \makebox[5em][c]{\textbf{76.60}\rlap{\tiny\,\ensuremath{\pm}\,0.39}} \\
\bottomrule
\end{tabular}}
\ifdim\wd\fgtabbox>\linewidth
  \resizebox{\linewidth}{!}{\usebox{\fgtabbox}}\else
  \usebox{\fgtabbox}\fi
\end{table}

\nointerlineskip\vskip 14pt
\begin{table}[!ht]
\caption{\textbf{Per-session accuracy on CUB-200.} Our results are averaged over three seeds. Acc./AA.: final/average accuracy; Base./Inc.: final base/new-class accuracy (\%). $\pm$: standard deviation. \textbf{Bold}/\underline{underline}: best/second among incremental methods; -: unavailable. Joint~Learning: trained on the training images of all sessions at once.}
\label{tab:supp:sessions_cub}
\centering
\setlength{\tabcolsep}{3.359pt}
\renewcommand{\arraystretch}{1.2}
\scriptsize
\settowidth{\hdw}{\footnotesize\textbf{Base.}}
\sbox{\fgtabbox}{\begin{tabular}{lllcccccccccccccc@{\hspace{1.3em}}}
\toprule
\multirow{2}{*}[-2.3pt]{\footnotesize\textbf{Method}} & \multirow{2}{*}[-2.3pt]{\footnotesize\textbf{Venue}} & \multirow{2}{*}[-2.3pt]{\footnotesize\textbf{Backbone}} & \multicolumn{10}{c}{\footnotesize\textbf{Sessions}} & \multicolumn{3}{c}{\footnotesize\textbf{Final session (10)}} & \multirow{2}{*}[-2.3pt]{\footnotesize\textbf{AA.}} \\
\cmidrule(lr){4-13} \cmidrule(lr){14-16}
 &  &  & \footnotesize\textbf{0} & \footnotesize\textbf{1} & \footnotesize\textbf{2} & \footnotesize\textbf{3} & \footnotesize\textbf{4} & \footnotesize\textbf{5} & \footnotesize\textbf{6} & \footnotesize\textbf{7} & \footnotesize\textbf{8} & \footnotesize\textbf{9} & \makebox[\hdw][c]{\footnotesize\textbf{Acc.}} & \makebox[\hdw][c]{\footnotesize\textbf{Base.}} & \makebox[\hdw][c]{\footnotesize\textbf{Inc.}} &  \\
\midrule
Joint Learning &  & \diagcell{ResNet-12 (scratch)} &  &  &  &  &  &  &  &  &  &  & 55.90 &  &  & \\
\midrule
\rowcolor{TableGroup}[\tabcolsep][\tabcolsep] \multicolumn{17}{l}{\textit{FSCIL methods}} \\
CEC~\cite{zhang2021CEC} & CVPR'21 & \diagcell{ResNet-12 (scratch)} & 59.94 & 54.31 & 49.88 & 45.81 & 44.44 & 41.72 & 39.95 & 37.74 & 36.79 & 35.74 & 34.50 & - & - & 43.71 \\
\rowrule
FACT~\cite{zhou2022FACT} & CVPR'22 & \diagcell{ResNet-12 (scratch)} & 58.45 & 53.14 & 48.77 & 45.05 & 43.97 & 40.57 & 38.77 & 36.62 & 35.60 & 33.86 & 32.71 & - & - & 42.50 \\
\rowrule
TEEN~\cite{wang2023teen} & NeurIPS'23 & \diagcell{ResNet-12 (scratch)} & 53.98 & 49.86 & 45.69 & 42.19 & 40.32 & 37.84 & 35.65 & 33.92 & 32.10 & 31.02 & 30.08 & - & - & 39.33 \\
\rowrule
CLOSER~\cite{oh2024CLOSER} & ECCV'24 & \diagcell{ResNet-12 (scratch)} & 67.35 & 55.90 & 51.37 & 47.37 & 44.22 & 41.22 & 38.86 & 36.94 & 35.04 & 33.46 & 31.74 & - & - & 43.95 \\
\rowrule
ADBS~\cite{li2025adbs} & AAAI'25 & \diagcell{ResNet-12 (scratch)} & 56.21 & 50.47 & 46.14 & 42.50 & 39.39 & 36.71 & 34.36 & 32.30 & 30.47 & 29.36 & 27.85 & - & - & 38.71 \\
\rowrule
DPCL~\cite{mao2026dpcl} & CVPRF'26 & \diagcell{ResNet-12 (scratch)} & 72.60 & 67.23 & 63.20 & 59.17 & 57.42 & 53.60 & 51.30 & 48.96 & 48.20 & 46.37 & 45.04 & 67.56 & 23.07 & 55.74 \\
\addlinespace[2pt]
\rowcolor{TableGroup}[\tabcolsep][\tabcolsep] \multicolumn{17}{l}{\textit{Fine-grained FSCIL methods}} \\
PFR~\cite{pan2024pfr} & PR'24 & \diagcell{ResNet-12 (scratch)} & 59.32 & 55.19 & 50.03 & 46.26 & 45.24 & 41.59 & 39.64 & 37.72 & 36.75 & 35.01 & 33.99 & - & - & 43.70 \\
\rowrule
MEL~\cite{zhao2025mel} & NeurIPS'25 & \diagcell{ResNet-12 (scratch)} & 60.40 & 56.43 & 52.00 & 47.88 & 46.69 & 43.34 & 41.65 & 39.96 & 38.07 & 37.37 & 36.00 & - & - & 45.44 \\
\rowrule
LGF~\cite{yin2026lgf} & IJMIR'26 & \diagcell{ResNet-12 (scratch)} & \underline{80.11} & \underline{76.38} & \underline{72.01} & \underline{67.22} & \underline{66.46} & \underline{62.42} & \underline{60.56} & \underline{58.44} & \underline{56.75} & \underline{55.88} & \underline{54.46} & 75.21 & 34.20 & \underline{64.61} \\
\addlinespace[2pt]
\rowcolor{TableHighlight}[\tabcolsep][1.3em] STAND (ours) &  & \diagcell{ResNet-12 (scratch)} & \textbf{83.75} & \textbf{79.92} & \textbf{77.60} & \textbf{73.33} & \textbf{73.38} & \textbf{70.84} & \textbf{69.88} & \textbf{68.70} & \textbf{67.04} & \textbf{66.79} & \textbf{65.57} & 78.17 & 53.29 & \makebox[5em][c]{\textbf{72.44}\rlap{\tiny\,\ensuremath{\pm}\,0.17}} \\
\bottomrule
\end{tabular}}
\ifdim\wd\fgtabbox>\linewidth
  \resizebox{\linewidth}{!}{\usebox{\fgtabbox}}\else
  \usebox{\fgtabbox}\fi
\end{table}
}
\wd0=\columnwidth
\box0
\endfgfloat
\makeatother
\endgroup

Tabs.~\ref{tab:supp:sessions_aircraft}--\ref{tab:supp:sessions_dogs} give the accuracy of each session behind Fig.~\ref{fig:teaser}, Tab.~\ref{tab:comparison}, and Fig.~\ref{fig:sessions} of the main paper, along with the final Base.\ and Inc. STAND (ours) achieves the highest final Acc.\ and AA.\ on all four datasets. Among the methods that report Base.\ and Inc., it also has the highest Base.\ on every dataset and the highest Inc.\ on FGVC-Aircraft, CUB-200, and Stanford Dogs. Moreover, its final Acc.\ exceeds that of Joint Learning by 9.67 to 20.87pp, although Joint Learning is trained on the training images of all sessions at once.

\onecolumn
\begingroup
\makeatletter
\@fptop=0pt
\@fpsep=14pt plus 2pt minus 1pt
\setlength{\intextsep}{8pt plus 2pt minus 2pt}
\makeatother
\begin{table}[!ht]
\caption{\textbf{Per-session accuracy on Stanford Cars.} Our results are averaged over three seeds. Acc./AA.: final/average accuracy; Base./Inc.: final base/new-class accuracy (\%). $\pm$: standard deviation. \textbf{Bold}/\underline{underline}: best/second among incremental methods; -: unavailable. Joint~Learning: trained on the training images of all sessions at once.}
\label{tab:supp:sessions_cars}
\centering
\setlength{\tabcolsep}{4.04pt}
\renewcommand{\arraystretch}{1.2}
\scriptsize
\settowidth{\hdw}{\footnotesize\textbf{Base.}}
\sbox{\fgtabbox}{\begin{tabular}{lllccccccccccccc@{\hspace{1.3em}}}
\toprule
\multirow{2}{*}[-2.3pt]{\footnotesize\textbf{Method}} & \multirow{2}{*}[-2.3pt]{\footnotesize\textbf{Venue}} & \multirow{2}{*}[-2.3pt]{\footnotesize\textbf{Backbone}} & \multicolumn{9}{c}{\footnotesize\textbf{Sessions}} & \multicolumn{3}{c}{\footnotesize\textbf{Final session (9)}} & \multirow{2}{*}[-2.3pt]{\footnotesize\textbf{AA.}} \\
\cmidrule(lr){4-12} \cmidrule(lr){13-15}
 &  &  & \footnotesize\textbf{0} & \footnotesize\textbf{1} & \footnotesize\textbf{2} & \footnotesize\textbf{3} & \footnotesize\textbf{4} & \footnotesize\textbf{5} & \footnotesize\textbf{6} & \footnotesize\textbf{7} & \footnotesize\textbf{8} & \makebox[\hdw][c]{\footnotesize\textbf{Acc.}} & \makebox[\hdw][c]{\footnotesize\textbf{Base.}} & \makebox[\hdw][c]{\footnotesize\textbf{Inc.}} &  \\
\midrule
Joint Learning &  & \diagcell{ResNet-12 (scratch)} &  &  &  &  &  &  &  &  &  & 58.84 &  &  & \\
\midrule
\rowcolor{TableGroup}[\tabcolsep][\tabcolsep] \multicolumn{16}{l}{\textit{FSCIL methods}} \\
CEC~\cite{zhang2021CEC} & CVPR'21 & \diagcell{ResNet-12 (scratch)} & 65.23 & 59.22 & 54.43 & 50.75 & 47.57 & 45.38 & 42.41 & 41.16 & 39.17 & 37.09 & - & - & 48.24 \\
\rowrule
FACT~\cite{zhou2022FACT} & CVPR'22 & \diagcell{ResNet-12 (scratch)} & 76.76 & 68.57 & 63.29 & 57.93 & 54.59 & 52.69 & 49.77 & 48.02 & 46.12 & 43.98 & - & - & 56.17 \\
\rowrule
SAVC~\cite{song2023SAVC} & CVPR'23 & \diagcell{ResNet-18 (IN-1K)} & - & - & - & - & - & - & - & - & - & 55.70 & 77.97 & 41.31 & 65.90 \\
\rowrule
TEEN~\cite{wang2023teen} & NeurIPS'23 & \diagcell{ResNet-12 (scratch)} & 75.42 & 69.39 & 64.30 & 60.24 & 56.44 & 54.32 & 50.96 & 49.78 & 47.56 & 45.32 & - & - & 57.37 \\
\rowrule
PriViLege~\cite{park2024PriViLege} & CVPR'24 & \diagcell{ViT-B/16 (IN-21K)} & - & - & - & - & - & - & - & - & - & 39.34 & 46.28 & 34.86 & 47.16 \\
\rowrule
CLOSER~\cite{oh2024CLOSER} & ECCV'24 & \diagcell{ResNet-12 (scratch)} & \underline{88.56} & 73.75 & 64.85 & 57.85 & 51.58 & 47.01 & 42.98 & 39.78 & 36.92 & 34.49 & - & - & 53.78 \\
\rowrule
ADBS~\cite{li2025adbs} & AAAI'25 & \diagcell{ResNet-12 (scratch)} & 75.91 & 64.64 & 56.16 & 49.65 & 44.49 & 40.79 & 37.24 & 34.26 & 32.02 & 29.79 & - & - & 46.50 \\
\rowrule
BiMC~\cite{chen2025BiMC} & CVPR'25 & \diagcell{ViT-B/16 (CLIP)} & - & - & - & - & - & - & - & - & - & \underline{74.11} & 78.72 & 71.13 & \underline{79.41} \\
\rowrule
HyCal~\cite{lee2026hycal} & CVPR'26 & \diagcell{ViT-B/16 (CLIP)} & - & - & - & - & - & - & - & - & - & 68.81 & 70.74 & 67.56 & 76.23 \\
\rowrule
DPCL~\cite{mao2026dpcl} & CVPRF'26 & \diagcell{ResNet-12 (scratch)} & 87.71 & \underline{82.07} & \underline{77.14} & \underline{72.65} & \underline{68.52} & \underline{66.82} & \underline{64.01} & \underline{61.65} & \underline{59.78} & 57.61 & 77.50 & 44.77 & 69.80 \\
\addlinespace[2pt]
\rowcolor{TableGroup}[\tabcolsep][\tabcolsep] \multicolumn{16}{l}{\textit{Fine-grained FSCIL methods}} \\
PFR~\cite{pan2024pfr} & PR'24 & \diagcell{ResNet-12 (scratch)} & 82.02 & 72.22 & 66.19 & 60.73 & 56.19 & 53.04 & 50.00 & 49.22 & 47.02 & 44.56 & - & - & 58.12 \\
\rowrule
PE-Net~\cite{nie2025penet} & ICONIP'25 & \diagcell{ResNet-12 (scratch)} & 82.09 & 73.53 & 67.74 & 62.97 & 58.34 & 56.62 & 53.88 & 50.96 & 48.88 & 46.18 & - & - & 60.12 \\
\rowrule
MEL~\cite{zhao2025mel} & NeurIPS'25 & \diagcell{ResNet-12 (scratch)} & 85.63 & 79.62 & 74.46 & 70.24 & 67.17 & 65.65 & 63.15 & 61.41 & 59.36 & 57.58 & - & - & 68.43 \\
\rowrule
LGF~\cite{yin2026lgf} & IJMIR'26 & \diagcell{ResNet-12 (scratch)} & 82.49 & 74.98 & 69.68 & 65.37 & 61.38 & 59.57 & 56.70 & 55.37 & 53.10 & 50.83 & - & - & 62.95 \\
\addlinespace[2pt]
\rowcolor{TableHighlight}[\tabcolsep][1.3em] STAND (ours) &  & \diagcell{ResNet-12 (scratch)} & \textbf{95.70} & \textbf{93.35} & \textbf{89.80} & \textbf{87.40} & \textbf{84.45} & \textbf{83.89} & \textbf{82.30} & \textbf{80.68} & \textbf{79.85} & \textbf{78.26} & 91.73 & 69.55 & \makebox[5em][c]{\textbf{85.57}\rlap{\tiny\,\ensuremath{\pm}\,0.33}} \\
\bottomrule
\end{tabular}}
\ifdim\wd\fgtabbox>\linewidth
  \resizebox{\linewidth}{!}{\usebox{\fgtabbox}}\else
  \usebox{\fgtabbox}\fi
\end{table}

\begin{table}[!ht]
\caption{\textbf{Per-session accuracy on Stanford Dogs.} Our results are averaged over three seeds. Acc./AA.: final/average accuracy; Base./Inc.: final base/new-class accuracy (\%). $\pm$: standard deviation. \textbf{Bold}/\underline{underline}: best/second among incremental methods; -: unavailable. Joint~Learning: trained on the training images of all sessions at once.}
\label{tab:supp:sessions_dogs}
\centering
\setlength{\tabcolsep}{4.89pt}
\renewcommand{\arraystretch}{1.2}
\scriptsize
\settowidth{\hdw}{\footnotesize\textbf{Base.}}
\sbox{\fgtabbox}{\begin{tabular}{lllcccccccccccc@{\hspace{1.3em}}}
\toprule
\multirow{2}{*}[-2.3pt]{\footnotesize\textbf{Method}} & \multirow{2}{*}[-2.3pt]{\footnotesize\textbf{Venue}} & \multirow{2}{*}[-2.3pt]{\footnotesize\textbf{Backbone}} & \multicolumn{8}{c}{\footnotesize\textbf{Sessions}} & \multicolumn{3}{c}{\footnotesize\textbf{Final session (8)}} & \multirow{2}{*}[-2.3pt]{\footnotesize\textbf{AA.}} \\
\cmidrule(lr){4-11} \cmidrule(lr){12-14}
 &  &  & \footnotesize\textbf{0} & \footnotesize\textbf{1} & \footnotesize\textbf{2} & \footnotesize\textbf{3} & \footnotesize\textbf{4} & \footnotesize\textbf{5} & \footnotesize\textbf{6} & \footnotesize\textbf{7} & \makebox[\hdw][c]{\footnotesize\textbf{Acc.}} & \makebox[\hdw][c]{\footnotesize\textbf{Base.}} & \makebox[\hdw][c]{\footnotesize\textbf{Inc.}} &  \\
\midrule
Joint Learning &  & \diagcell{ResNet-12 (scratch)} &  &  &  &  &  &  &  &  & 35.76 &  &  & \\
\midrule
\rowcolor{TableGroup}[\tabcolsep][\tabcolsep] \multicolumn{15}{l}{\textit{FSCIL methods}} \\
CEC~\cite{zhang2021CEC} & CVPR'21 & \diagcell{ResNet-12 (scratch)} & 55.01 & 46.05 & 43.83 & 41.57 & 39.28 & 37.26 & 35.53 & 33.87 & 32.65 & - & - & 40.56 \\
\rowrule
FACT~\cite{zhou2022FACT} & CVPR'22 & \diagcell{ResNet-12 (scratch)} & 58.43 & 50.92 & 47.48 & 44.39 & 41.48 & 39.04 & 36.78 & 34.87 & 33.26 & - & - & 42.96 \\
\rowrule
TEEN~\cite{wang2023teen} & NeurIPS'23 & \diagcell{ResNet-12 (scratch)} & 61.09 & 54.70 & 51.71 & 49.56 & 47.28 & 44.41 & 42.30 & 40.56 & 39.14 & - & - & 47.86 \\
\rowrule
CLOSER~\cite{oh2024CLOSER} & ECCV'24 & \diagcell{ResNet-12 (scratch)} & 60.22 & 53.22 & 49.40 & 46.64 & 43.24 & 40.65 & 38.33 & 36.10 & 34.37 & - & - & 44.69 \\
\rowrule
ADBS~\cite{li2025adbs} & AAAI'25 & \diagcell{ResNet-12 (scratch)} & 61.63 & 53.15 & 47.55 & 43.02 & 39.28 & 35.78 & 33.16 & 31.16 & 29.15 & - & - & 41.54 \\
\rowrule
DPCL~\cite{mao2026dpcl} & CVPRF'26 & \diagcell{ResNet-12 (scratch)} & 58.97 & 50.25 & 47.46 & 44.62 & 41.89 & 40.25 & 37.94 & 35.94 & 34.50 & 46.41 & 23.82 & 43.54 \\
\addlinespace[2pt]
\rowcolor{TableGroup}[\tabcolsep][\tabcolsep] \multicolumn{15}{l}{\textit{Fine-grained FSCIL methods}} \\
PFR~\cite{pan2024pfr} & PR'24 & \diagcell{ResNet-12 (scratch)} & 59.78 & 51.65 & 48.14 & 45.91 & 41.13 & 40.13 & 38.20 & 37.28 & 36.01 & - & - & 44.25 \\
\rowrule
PE-Net~\cite{nie2025penet} & ICONIP'25 & \diagcell{ResNet-12 (scratch)} & 60.94 & 54.16 & 51.28 & 49.21 & 46.16 & 43.67 & 41.84 & 40.04 & 38.76 & - & - & 47.34 \\
\rowrule
MEL~\cite{zhao2025mel} & NeurIPS'25 & \diagcell{ResNet-12 (scratch)} & 65.57 & 58.44 & 56.03 & 53.84 & \underline{51.34} & \underline{49.45} & \underline{47.42} & \underline{45.29} & \underline{44.06} & - & - & \underline{52.38} \\
\rowrule
LGF~\cite{yin2026lgf} & IJMIR'26 & \diagcell{ResNet-12 (scratch)} & \underline{68.47} & \underline{59.15} & \underline{56.20} & \underline{54.01} & 50.89 & 48.56 & 46.37 & 44.27 & 42.74 & - & - & 52.30 \\
\addlinespace[2pt]
\rowcolor{TableHighlight}[\tabcolsep][1.3em] STAND (ours) &  & \diagcell{ResNet-12 (scratch)} & \textbf{74.81} & \textbf{68.25} & \textbf{65.35} & \textbf{63.39} & \textbf{61.17} & \textbf{58.94} & \textbf{56.55} & \textbf{54.50} & \textbf{52.76} & 70.40 & 36.86 & \makebox[5em][c]{\textbf{61.74}\rlap{\tiny\,\ensuremath{\pm}\,0.19}} \\
\bottomrule
\end{tabular}}
\ifdim\wd\fgtabbox>\linewidth
  \resizebox{\linewidth}{!}{\usebox{\fgtabbox}}\else
  \usebox{\fgtabbox}\fi
\end{table}

\clearpage
\endgroup
\twocolumn


{\small
\begin{thebibliography}{69}
\providecommand{\natexlab}[1]{#1}
\providecommand{\url}[1]{\texttt{#1}}
\expandafter\ifx\csname urlstyle\endcsname\relax
  \providecommand{\doi}[1]{doi: #1}\else
  \providecommand{\doi}{doi: \begingroup \urlstyle{rm}\Url}\fi

\bibitem[Rebuffi et~al.(2017)Rebuffi, Kolesnikov, Sperl, and
  Lampert]{rebuffi2017icarl}
Sylvestre-Alvise Rebuffi, Alexander Kolesnikov, Georg Sperl, and Christoph~H.
  Lampert.
\newblock {iCaRL}: Incremental classifier and representation learning.
\newblock In \emph{CVPR}, 2017.

\bibitem[Petit et~al.(2023)Petit, Popescu, Schindler, Picard, and
  Delezoide]{petit2023fetril}
Gr{\'e}goire Petit, Adrian Popescu, Hugo Schindler, David Picard, and Bertrand
  Delezoide.
\newblock {FeTrIL}: Feature translation for exemplar-free class-incremental
  learning.
\newblock In \emph{WACV}, 2023.

\bibitem[Magistri et~al.(2024)Magistri, Trinci, Soutif-Cormerais, van~de
  Weijer, and Bagdanov]{magistri2024efc}
Simone Magistri, Tomaso Trinci, Albin Soutif-Cormerais, Joost van~de Weijer,
  and Andrew~D. Bagdanov.
\newblock Elastic feature consolidation for cold start exemplar-free
  incremental learning.
\newblock In \emph{ICLR}, 2024.

\bibitem[Tao et~al.(2020)Tao, Hong, Chang, Dong, Wei, and Gong]{tao2020TOPIC}
Xiaoyu Tao, Xiaopeng Hong, Xinyuan Chang, Songlin Dong, Xing Wei, and Yihong
  Gong.
\newblock Few-shot class-incremental learning.
\newblock In \emph{CVPR}, 2020.

\bibitem[Zhang et~al.(2021)Zhang, Song, Lin, Zheng, Pan, and Xu]{zhang2021CEC}
Chi Zhang, Nan Song, Guosheng Lin, Yun Zheng, Pan Pan, and Yinghui Xu.
\newblock Few-shot incremental learning with continually evolved classifiers.
\newblock In \emph{CVPR}, 2021.

\bibitem[Zhou et~al.(2022)Zhou, Wang, Ye, Ma, Pu, and Zhan]{zhou2022FACT}
Da-Wei Zhou, Fu-Yun Wang, Han-Jia Ye, Liang Ma, Shiliang Pu, and De-Chuan Zhan.
\newblock Forward compatible few-shot class-incremental learning.
\newblock In \emph{CVPR}, 2022.

\bibitem[Vinyals et~al.(2016)Vinyals, Blundell, Lillicrap, Kavukcuoglu, and
  Wierstra]{vinyals2016matching_mini}
Oriol Vinyals, Charles Blundell, Timothy Lillicrap, Koray Kavukcuoglu, and Daan
  Wierstra.
\newblock Matching networks for one shot learning.
\newblock In \emph{NeurIPS}, 2016.

\bibitem[Krizhevsky(2009)]{krizhevsky2009CIFAR100}
Alex Krizhevsky.
\newblock Learning multiple layers of features from tiny images.
\newblock Technical report, University of Toronto, 2009.

\bibitem[Pan et~al.(2024)Pan, Zhang, Yu, Zhang, and Gao]{pan2024pfr}
Zicheng Pan, Weichuan Zhang, Xiaohan Yu, Miaohua Zhang, and Yongsheng Gao.
\newblock Pseudo-set frequency refinement architecture for fine-grained
  few-shot class-incremental learning.
\newblock \emph{Pattern Recognition}, 2024.

\bibitem[Zhao et~al.(2025)Zhao, Chen, Wang, Luo, and Xu]{zhao2025mel}
Li-Jun Zhao, Zhen-Duo Chen, Yongxin Wang, Xin Luo, and Xin-Shun Xu.
\newblock Evolving and regularizing meta-environment learner for fine-grained
  few-shot class-incremental learning.
\newblock In \emph{NeurIPS}, 2025.

\bibitem[Yin et~al.(2026)Yin, Liu, Xu, Zhang, and Zhang]{yin2026lgf}
Yimin Yin, Haoling Liu, Sihang Xu, Renye Zhang, and Jinghua Zhang.
\newblock Optimising few-shot class-incremental learning for fine-grained
  visual recognition.
\newblock \emph{IJMIR}, 2026.

\bibitem[Nie et~al.(2025)Nie, Pan, and Gao]{nie2025penet}
Hongyu Nie, Zicheng Pan, and Yongsheng Gao.
\newblock Purification enhancement network for fine-grained few-shot class
  incremental learning.
\newblock In \emph{ICONIP}, 2025.

\bibitem[Song et~al.(2023)Song, Zhao, Shi, Peng, Yuan, and Tian]{song2023SAVC}
Zeyin Song, Yifan Zhao, Yujun Shi, Peixi Peng, Li Yuan, and Yonghong Tian.
\newblock Learning with fantasy: Semantic-aware virtual contrastive constraint
  for few-shot class-incremental learning.
\newblock In \emph{CVPR}, 2023.

\bibitem[Park et~al.(2024)Park, Song, and Park]{park2024PriViLege}
Keon-Hee Park, Kyungwoo Song, and Gyeong-Moon Park.
\newblock Pre-trained vision and language transformers are few-shot incremental
  learners.
\newblock In \emph{CVPR}, 2024.

\bibitem[Chen et~al.(2025)Chen, Ding, Wang, Huo, Gao, and Li]{chen2025BiMC}
Yiyang Chen, Tianyu Ding, Lei Wang, Jing Huo, Yang Gao, and Wenbin Li.
\newblock Enhancing few-shot class-incremental learning via training-free
  bi-level modality calibration.
\newblock In \emph{CVPR}, 2025.

\bibitem[Lee et~al.(2026)Lee, Kim, Kwon, Lee, Kim, Jang, and Kim]{lee2026hycal}
Eunju Lee, MiHyeon Kim, JuneHyoung Kwon, Yoonji Lee, JiHyun Kim, Soojin Jang,
  and YoungBin Kim.
\newblock {HyCal}: A training-free prototype calibration method for
  cross-discipline few-shot class-incremental learning.
\newblock In \emph{CVPR}, 2026.

\bibitem[Radford et~al.(2021)Radford, Kim, Hallacy, Ramesh, Goh, Agarwal,
  Sastry, Askell, Mishkin, Clark, Krueger, and Sutskever]{radford2021CLIP}
Alec Radford, Jong~Wook Kim, Chris Hallacy, Aditya Ramesh, Gabriel Goh,
  Sandhini Agarwal, Girish Sastry, Amanda Askell, Pamela Mishkin, Jack Clark,
  Gretchen Krueger, and Ilya Sutskever.
\newblock Learning transferable visual models from natural language
  supervision.
\newblock In \emph{ICML}, 2021.

\bibitem[Hinton et~al.(2015)Hinton, Vinyals, and Dean]{hinton2015distilling}
Geoffrey Hinton, Oriol Vinyals, and Jeff Dean.
\newblock Distilling the knowledge in a neural network.
\newblock arXiv, 2015.

\bibitem[Oquab et~al.(2024)Oquab, Darcet, Moutakanni, Vo, Szafraniec, Khalidov,
  Fernandez, Haziza, Massa, El-Nouby, Assran, Ballas, Galuba, Howes, Huang, Li,
  Misra, Rabbat, Sharma, Synnaeve, Xu, Jegou, Mairal, Labatut, Joulin, and
  Bojanowski]{oquab2023DINOv2}
Maxime Oquab, Timoth{\'e}e Darcet, Th{\'e}o Moutakanni, Huy~V. Vo, Marc
  Szafraniec, Vasil Khalidov, Pierre Fernandez, Daniel Haziza, Francisco Massa,
  Alaaeldin El-Nouby, Mido Assran, Nicolas Ballas, Wojciech Galuba, Russell
  Howes, Po-Yao Huang, Shang-Wen Li, Ishan Misra, Michael Rabbat, Vasu Sharma,
  Gabriel Synnaeve, Hu Xu, Herv{\'e} Jegou, Julien Mairal, Patrick Labatut,
  Armand Joulin, and Piotr Bojanowski.
\newblock {DINO}v2: Learning robust visual features without supervision.
\newblock \emph{TMLR}, 2024.

\bibitem[Sim{\'e}oni et~al.(2026)Sim{\'e}oni, Vo, Seitzer, Baldassarre, Oquab,
  Jose, Khalidov, Szafraniec, Yi, Ramamonjisoa, Massa, Haziza, Wehrstedt, Wang,
  Darcet, Moutakanni, Sentana, Roberts, Vedaldi, Tolan, Brandt, Couprie,
  Mairal, J{\'e}gou, Labatut, and Bojanowski]{simeoni2025dinov3}
Oriane Sim{\'e}oni, Huy~V. Vo, Maximilian Seitzer, Federico Baldassarre, Maxime
  Oquab, Cijo Jose, Vasil Khalidov, Marc Szafraniec, Seung~Eun Yi, Micha{\"e}l
  Ramamonjisoa, Francisco Massa, Daniel Haziza, Luca Wehrstedt, Jianyuan Wang,
  Timoth{\'e}e Darcet, Th{\'e}o Moutakanni, Leonel Sentana, Claire Roberts,
  Andrea Vedaldi, Jamie Tolan, John Brandt, Camille Couprie, Julien Mairal,
  Herv{\'e} J{\'e}gou, Patrick Labatut, and Piotr Bojanowski.
\newblock {DINO}v3.
\newblock \emph{TMLR}, 2026.

\bibitem[Jeong et~al.(2026)Jeong, Lee, and Choi]{jeong2026skd}
Hyeon-Jin Jeong, Han-Jin Lee, and Seok-Hwan Choi.
\newblock Streamlined knowledge distillation.
\newblock In \emph{CVPR}, 2026.

\bibitem[Xu et~al.(2026)Xu, Zheng, Wang, Xu, and Cheng]{xu2026hcd}
Liuchi Xu, Hao Zheng, Lu Wang, Lisheng Xu, and Jun Cheng.
\newblock Heterogeneous complementary distillation.
\newblock In \emph{AAAI}, 2026.

\bibitem[Beyer et~al.(2022)Beyer, Zhai, Royer, Markeeva, Anil, and
  Kolesnikov]{beyer2022knowledge}
Lucas Beyer, Xiaohua Zhai, Am{\'e}lie Royer, Larisa Markeeva, Rohan Anil, and
  Alexander Kolesnikov.
\newblock Knowledge distillation: A good teacher is patient and consistent.
\newblock In \emph{CVPR}, 2022.

\bibitem[Zhang et~al.(2018)Zhang, Cisse, Dauphin, and
  Lopez-Paz]{zhang2018mixup}
Hongyi Zhang, Moustapha Cisse, Yann~N. Dauphin, and David Lopez-Paz.
\newblock mixup: Beyond empirical risk minimization.
\newblock In \emph{ICLR}, 2018.

\bibitem[Wang et~al.(2023)Wang, Zhou, Zhang, Zhan, and Ye]{wang2023teen}
Qi-Wei Wang, Da-Wei Zhou, Yi-Kai Zhang, De-Chuan Zhan, and Han-Jia Ye.
\newblock Few-shot class-incremental learning via training-free prototype
  calibration.
\newblock In \emph{NeurIPS}, 2023.

\bibitem[Oh et~al.(2024)Oh, Baik, and Lee]{oh2024CLOSER}
Junghun Oh, Sungyong Baik, and Kyoung~Mu Lee.
\newblock {CLOSER}: Towards better representation learning for few-shot
  class-incremental learning.
\newblock In \emph{ECCV}, 2024.

\bibitem[Li et~al.(2025)Li, Tan, Yang, Cheng, Dong, and Yang]{li2025adbs}
Linhao Li, Yongzhang Tan, Siyuan Yang, Hao Cheng, Yongfeng Dong, and Liang
  Yang.
\newblock Adaptive decision boundary for few-shot class-incremental learning.
\newblock In \emph{AAAI}, 2025.

\bibitem[Kim et~al.(2024)Kim, Hong, Kim, Kim, Tiruneh, On, Song, Choi, and
  Baek]{kim2024classwise}
Junsu Kim, Sumin Hong, Chanwoo Kim, Jihyeon Kim, Yihalem~Yimolal Tiruneh,
  Jeongwan On, Jihyun Song, Sunhwa Choi, and Seungryul Baek.
\newblock Class-wise buffer management for incremental object detection: An
  effective buffer training strategy.
\newblock In \emph{ICASSP}, 2024.

\bibitem[Mao et~al.(2026)Mao, Zhang, Li, Liu, Wu, and Wang]{mao2026dpcl}
Zhilong Mao, Hang Zhang, Yanmin Li, Lihua Liu, Jibing Wu, and Mao Wang.
\newblock Dynamic pseudo-label assignment and consistent prototypical learning
  for few-shot class-incremental learning.
\newblock In \emph{CVPR Findings}, 2026.

\bibitem[Yang et~al.(2023)Yang, Yuan, Li, Lin, Torr, and Tao]{yang2023NC_FSCIL}
Yibo Yang, Haobo Yuan, Xiangtai Li, Zhouchen Lin, Philip Torr, and Dacheng Tao.
\newblock Neural collapse inspired feature-classifier alignment for few-shot
  class-incremental learning.
\newblock In \emph{ICLR}, 2023.

\bibitem[Kim and Baek(2026)]{kim2026bee}
Junsu Kim and Seungryul Baek.
\newblock Where is the bee? {D}etecting tiny pollinators with a single
  collaborative-head transformer.
\newblock arXiv, 2026.

\bibitem[Tschannen et~al.(2025)Tschannen, Gritsenko, Wang, Naeem,
  Alabdulmohsin, Parthasarathy, Evans, Beyer, Xia, Mustafa, H{\'e}naff,
  Harmsen, Steiner, and Zhai]{tschannen2025siglip2}
Michael Tschannen, Alexey Gritsenko, Xiao Wang, Muhammad~Ferjad Naeem, Ibrahim
  Alabdulmohsin, Nikhil Parthasarathy, Talfan Evans, Lucas Beyer, Ye Xia, Basil
  Mustafa, Olivier H{\'e}naff, Jeremiah Harmsen, Andreas Steiner, and Xiaohua
  Zhai.
\newblock {SigLIP 2}: Multilingual vision-language encoders with improved
  semantic understanding, localization, and dense features.
\newblock arXiv, 2025.

\bibitem[Dosovitskiy et~al.(2021)Dosovitskiy, Beyer, Kolesnikov, Weissenborn,
  Zhai, Unterthiner, Dehghani, Minderer, Heigold, Gelly, Uszkoreit, and
  Houlsby]{dosovitskiy2021vit}
Alexey Dosovitskiy, Lucas Beyer, Alexander Kolesnikov, Dirk Weissenborn,
  Xiaohua Zhai, Thomas Unterthiner, Mostafa Dehghani, Matthias Minderer, Georg
  Heigold, Sylvain Gelly, Jakob Uszkoreit, and Neil Houlsby.
\newblock An image is worth 16x16 words: Transformers for image recognition at
  scale.
\newblock In \emph{ICLR}, 2021.

\bibitem[Rombach et~al.(2022)Rombach, Blattmann, Lorenz, Esser, and
  Ommer]{Rombach_2022LDM}
Robin Rombach, Andreas Blattmann, Dominik Lorenz, Patrick Esser, and Bj\"orn
  Ommer.
\newblock High-resolution image synthesis with latent diffusion models.
\newblock In \emph{CVPR}, 2022.

\bibitem[Esser et~al.(2024)Esser, Kulal, Blattmann, Entezari, M{\"u}ller,
  Saini, Levi, Lorenz, Sauer, Boesel, Podell, Dockhorn, English, and
  Rombach]{esser2024sd3}
Patrick Esser, Sumith Kulal, Andreas Blattmann, Rahim Entezari, Jonas
  M{\"u}ller, Harry Saini, Yam Levi, Dominik Lorenz, Axel Sauer, Frederic
  Boesel, Dustin Podell, Tim Dockhorn, Zion English, and Robin Rombach.
\newblock Scaling rectified flow transformers for high-resolution image
  synthesis.
\newblock In \emph{ICML}, 2024.

\bibitem[Kim et~al.(2024)Kim, Cho, Kim, Tiruneh, and Baek]{kim2024sddgr}
Junsu Kim, Hoseong Cho, Jihyeon Kim, Yihalem~Yimolal Tiruneh, and Seungryul
  Baek.
\newblock {SDDGR}: Stable diffusion-based deep generative replay for class
  incremental object detection.
\newblock In \emph{CVPR}, 2024.

\bibitem[Kim et~al.(2025)Kim, Ku, and Baek]{kim2025can}
Junsu Kim, Yunhoe Ku, and Seungryul Baek.
\newblock Can synthetic images conquer forgetting? {B}eyond unexplored doubts
  in few-shot class-incremental learning.
\newblock In \emph{ICCVW}, 2025.

\bibitem[Wu et~al.(2025)Wu, Zhao, Chen, and Li]{wu2025dcs}
Ruitao Wu, Yifan Zhao, Guangyao Chen, and Jia Li.
\newblock Diffusion-classifier synergy: Reward-aligned learning via mutual
  boosting loop for {FSCIL}.
\newblock In \emph{NeurIPS}, 2025.

\bibitem[Kim et~al.(2025)Kim, Ku, Han, and Baek]{kim2025beyond}
Junsu Kim, Yunhoe Ku, Dongyoon Han, and Seungryul Baek.
\newblock Beyond synthetic replays: Turning diffusion features into few-shot
  class-incremental learning knowledge.
\newblock arXiv, 2025.

\bibitem[Kim et~al.(2024)Kim, Ku, Kim, Cha, and Baek]{kim2024vlmpl}
Junsu Kim, Yunhoe Ku, Jihyeon Kim, Junuk Cha, and Seungryul Baek.
\newblock {VLM-PL}: Advanced pseudo labeling approach for class incremental
  object detection via vision-language model.
\newblock In \emph{CVPRW}, 2024.

\bibitem[Sar{\i}y{\i}ld{\i}z et~al.(2024)Sar{\i}y{\i}ld{\i}z, Weinzaepfel,
  Lucas, Larlus, and Kalantidis]{sariyildiz2024unic}
Mert~B{\"u}lent Sar{\i}y{\i}ld{\i}z, Philippe Weinzaepfel, Thomas Lucas, Diane
  Larlus, and Yannis Kalantidis.
\newblock {UNIC}: Universal classification models via multi-teacher
  distillation.
\newblock In \emph{ECCV}, 2024.

\bibitem[Chen et~al.(2022)Chen, Mei, Zhang, Wang, Feng, and
  Chen]{chen2022simkd}
Defang Chen, Jian-Ping Mei, Hailin Zhang, Can Wang, Yan Feng, and Chun Chen.
\newblock Knowledge distillation with the reused teacher classifier.
\newblock In \emph{CVPR}, 2022.

\bibitem[Marrie et~al.(2024)Marrie, Arbel, Mairal, and Larlus]{marrie2024good}
Juliette Marrie, Michael Arbel, Julien Mairal, and Diane Larlus.
\newblock On good practices for task-specific distillation of large pretrained
  visual models.
\newblock \emph{TMLR}, 2024.

\bibitem[Zhao et~al.(2022)Zhao, Cui, Song, Qiu, and Liang]{zhao2022dkd}
Borui Zhao, Quan Cui, Renjie Song, Yiyu Qiu, and Jiajun Liang.
\newblock Decoupled knowledge distillation.
\newblock In \emph{CVPR}, 2022.

\bibitem[Huang et~al.(2022)Huang, You, Wang, Qian, and Xu]{huang2022dist}
Tao Huang, Shan You, Fei Wang, Chen Qian, and Chang Xu.
\newblock Knowledge distillation from a stronger teacher.
\newblock In \emph{NeurIPS}, 2022.

\bibitem[Romero et~al.(2015)Romero, Ballas, Kahou, Chassang, Gatta, and
  Bengio]{romero2015fitnets}
Adriana Romero, Nicolas Ballas, Samira~Ebrahimi Kahou, Antoine Chassang, Carlo
  Gatta, and Yoshua Bengio.
\newblock {FitNets}: Hints for thin deep nets.
\newblock In \emph{ICLR}, 2015.

\bibitem[Park et~al.(2019)Park, Kim, Lu, and Cho]{park2019rkd}
Wonpyo Park, Dongju Kim, Yan Lu, and Minsu Cho.
\newblock Relational knowledge distillation.
\newblock In \emph{CVPR}, 2019.

\bibitem[Passalis and Tefas(2018)]{passalis2018pkt}
Nikolaos Passalis and Anastasios Tefas.
\newblock Learning deep representations with probabilistic knowledge transfer.
\newblock In \emph{ECCV}, 2018.

\bibitem[Koohpayegani et~al.(2020)Koohpayegani, Tejankar, and
  Pirsiavash]{koohpayegani2020compress}
Soroush~Abbasi Koohpayegani, Ajinkya Tejankar, and Hamed Pirsiavash.
\newblock {CompRess}: Self-supervised learning by compressing representations.
\newblock In \emph{NeurIPS}, 2020.

\bibitem[Fang et~al.(2021)Fang, Wang, Wang, Zhang, Yang, and Liu]{fang2021seed}
Zhiyuan Fang, Jianfeng Wang, Lijuan Wang, Lei Zhang, Yezhou Yang, and Zicheng
  Liu.
\newblock {SEED}: Self-supervised distillation for visual representation.
\newblock In \emph{ICLR}, 2021.

\bibitem[Deng et~al.(2009)Deng, Dong, Socher, Li, Li, and
  Fei-Fei]{deng2009imagenet}
Jia Deng, Wei Dong, Richard Socher, Li-Jia Li, Kai Li, and Li Fei-Fei.
\newblock {ImageNet}: A large-scale hierarchical image database.
\newblock In \emph{CVPR}, 2009.

\bibitem[Bolya et~al.(2025)Bolya, Huang, Sun, Cho, Madotto, Wei, Ma, Zhi,
  Rajasegaran, Bangalath, Wang, Monteiro, Xu, Dong, Ravi, Li, Dollar, and
  Feichtenhofer]{bolya2025perception}
Daniel Bolya, Po-Yao Huang, Peize Sun, Jang~Hyun Cho, Andrea Madotto, Chen Wei,
  Tengyu Ma, Jiale Zhi, Jathushan Rajasegaran, Hanoona Bangalath, Junke Wang,
  Marco Monteiro, Hu Xu, Shiyu Dong, Nikhila Ravi, Shang-Wen Li, Piotr Dollar,
  and Christoph Feichtenhofer.
\newblock Perception encoder: The best visual embeddings are not at the output
  of the network.
\newblock In \emph{NeurIPS}, 2025.

\bibitem[Ranzinger et~al.(2026)Ranzinger, Heinrich, McCarthy, Kautz, Tao,
  Catanzaro, and Molchanov]{ranzinger2026cradio}
Mike Ranzinger, Greg Heinrich, Collin McCarthy, Jan Kautz, Andrew Tao, Bryan
  Catanzaro, and Pavlo Molchanov.
\newblock {C-RADIOv4} (tech report).
\newblock arXiv, 2026.

\bibitem[Fini et~al.(2025)Fini, Shukor, Li, Dufter, Klein, Haldimann,
  Aitharaju, Turrisi~da Costa, B{\'e}thune, Gan, Toshev, Eichner, Nabi, Yang,
  Susskind, and El-Nouby]{fini2025aimv2}
Enrico Fini, Mustafa Shukor, Xiujun Li, Philipp Dufter, Michal Klein, David
  Haldimann, Sai Aitharaju, Victor~G. Turrisi~da Costa, Louis B{\'e}thune, Zhe
  Gan, Alexander Toshev, Marcin Eichner, Moin Nabi, Yinfei Yang, Joshua
  Susskind, and Alaaeldin El-Nouby.
\newblock Multimodal autoregressive pre-training of large vision encoders.
\newblock In \emph{CVPR}, 2025.

\bibitem[Liu et~al.(2024)Liu, Chen, and Liu]{liu2024dccd}
Yuang Liu, Jun Chen, and Yong Liu.
\newblock {DCCD}: Reducing neural network redundancy via distillation.
\newblock \emph{IEEE TNNLS}, 2024.

\bibitem[Maji et~al.(2013)Maji, Rahtu, Kannala, Blaschko, and
  Vedaldi]{maji2013fgvc}
Subhransu Maji, Esa Rahtu, Juho Kannala, Matthew Blaschko, and Andrea Vedaldi.
\newblock Fine-grained visual classification of aircraft.
\newblock arXiv, 2013.

\bibitem[Krause et~al.(2013)Krause, Stark, Deng, and Fei-Fei]{krause2013cars}
Jonathan Krause, Michael Stark, Jia Deng, and Li Fei-Fei.
\newblock {3D} object representations for fine-grained categorization.
\newblock In \emph{ICCVW}, 2013.

\bibitem[Wah et~al.(2011)Wah, Branson, Welinder, Perona, and
  Belongie]{wah2011CUB200}
Catherine Wah, Steve Branson, Peter Welinder, Pietro Perona, and Serge
  Belongie.
\newblock The {Caltech-UCSD} birds-200-2011 dataset.
\newblock Technical report, California Institute of Technology, 2011.

\bibitem[Khosla et~al.(2011)Khosla, Jayadevaprakash, Yao, and
  Li]{khosla2011dogs}
Aditya Khosla, Nityananda Jayadevaprakash, Bangpeng Yao, and Fei-Fei Li.
\newblock Novel dataset for fine-grained image categorization: {S}tanford dogs.
\newblock In \emph{CVPRW}, 2011.

\bibitem[Yang et~al.(2023)Yang, Li, Zhao, Cai, Zhang, Ma, Zhao, Chen, Jiang,
  Sun, Teng, Qi, Huang, and Grzegorzek]{yang2023emds7}
Hechen Yang, Chen Li, Xin Zhao, Bencheng Cai, Jiawei Zhang, Pingli Ma, Peng
  Zhao, Ao Chen, Tao Jiang, Hongzan Sun, Yueyang Teng, Shouliang Qi, Xinyu
  Huang, and Marcin Grzegorzek.
\newblock {EMDS-7}: Environmental microorganism image dataset seventh version
  for multiple object detection evaluation.
\newblock \emph{Frontiers in Microbiology}, 2023.

\bibitem[Zhou et~al.(2026)Zhou, Zhang, Xu, Hu, Zeng, and Yin]{zhou2026emds7}
Jinyi Zhou, Yinuo Zhang, Sihang Xu, Yangfan Hu, Jinhui Zeng, and Yimin Yin.
\newblock {EMDS-7-FSCIL}: A benchmark for few-shot class-incremental learning
  in environmental microorganism recognition.
\newblock \emph{Frontiers in Microbiology}, 2026.

\bibitem[Yun et~al.(2019)Yun, Han, Oh, Chun, Choe, and Yoo]{yun2019cutmix}
Sangdoo Yun, Dongyoon Han, Seong~Joon Oh, Sanghyuk Chun, Junsuk Choe, and
  Youngjoon Yoo.
\newblock {CutMix}: Regularization strategy to train strong classifiers with
  localizable features.
\newblock In \emph{ICCV}, 2019.

\bibitem[He et~al.(2016)He, Zhang, Ren, and Sun]{he2016resnet}
Kaiming He, Xiangyu Zhang, Shaoqing Ren, and Jian Sun.
\newblock Deep residual learning for image recognition.
\newblock In \emph{CVPR}, 2016.

\bibitem[Cubuk et~al.(2019)Cubuk, Zoph, Mane, Vasudevan, and
  Le]{cubuk2019autoaugment}
Ekin~D. Cubuk, Barret Zoph, Dandelion Mane, Vijay Vasudevan, and Quoc~V. Le.
\newblock {AutoAugment}: Learning augmentation strategies from data.
\newblock In \emph{CVPR}, 2019.

\bibitem[Cho and Hariharan(2019)]{cho2019efficacy}
Jang~Hyun Cho and Bharath Hariharan.
\newblock On the efficacy of knowledge distillation.
\newblock In \emph{ICCV}, 2019.

\bibitem[Mirzadeh et~al.(2020)Mirzadeh, Farajtabar, Li, Levine, Matsukawa, and
  Ghasemzadeh]{mirzadeh2020improved}
Seyed~Iman Mirzadeh, Mehrdad Farajtabar, Ang Li, Nir Levine, Akihiro Matsukawa,
  and Hassan Ghasemzadeh.
\newblock Improved knowledge distillation via teacher assistant.
\newblock In \emph{AAAI}, 2020.

\bibitem[Loshchilov and Hutter(2019)]{loshchilov2017AdamW}
Ilya Loshchilov and Frank Hutter.
\newblock Decoupled weight decay regularization.
\newblock In \emph{ICLR}, 2019.

\bibitem[Qin et~al.(2020)Qin, Fang, Zhang, Liu, Wang, and
  Wang]{qin2020resizemix}
Jie Qin, Jiemin Fang, Qian Zhang, Wenyu Liu, Xingang Wang, and Xinggang Wang.
\newblock {ResizeMix}: Mixing data with preserved object information and true
  labels.
\newblock arXiv, 2020.

\bibitem[Sandler et~al.(2018)Sandler, Howard, Zhu, Zhmoginov, and
  Chen]{sandler2018mobilenetv2}
Mark Sandler, Andrew Howard, Menglong Zhu, Andrey Zhmoginov, and Liang-Chieh
  Chen.
\newblock {MobileNetV2}: Inverted residuals and linear bottlenecks.
\newblock In \emph{CVPR}, 2018.

\end{thebibliography}

}
\end{document}